\documentclass[sigconf]{acmart}
\usepackage{multirow}
\usepackage{enumitem}
\usepackage{subcaption}
\usepackage{bbm}
\usepackage{graphicx}
\usepackage{mathtools}

\usepackage[linesnumbered,ruled,vlined]{algorithm2e}
\AtBeginDocument{%
  \providecommand\BibTeX{{%
    \normalfont B\kern-0.5em{\scshape i\kern-0.25em b}\kern-0.8em\TeX}}}

\setcopyright{acmcopyright}
\copyrightyear{2022}
\acmYear{2022}
\setcopyright{acmcopyright}\acmConference[WWW '22]{Proceedings of the ACM Web Conference 2022}{April 25--29, 2022}{Virtual Event, Lyon, France}
\acmBooktitle{Proceedings of the ACM Web Conference 2022 (WWW '22), April 25--29, 2022, Virtual Event, Lyon, France}
\acmPrice{15.00}
\acmDOI{10.1145/3485447.3512090}
\acmISBN{978-1-4503-9096-5/22/04}

\newcommand{\lpmname}{\textsf{ICL}\xspace}

\newcommand{\modelwithtansformer}{\textsf{ICLRec}\xspace}

\setlist[itemize]{leftmargin=*}
 
\begin{document}

\title{Intent Contrastive Learning for Sequential Recommendation}

% \author[1]{Yongjun Chen}
% \author[1]{Zhiwei Liu}
% \author[1]{Jia Li}
% \author[2]{Julian McAuley}
% \author[1]{Caiming Xiong}
% \affil[1]{Salesforce Research, Palo Alto, CA, USA}
% \affil[2]{University of California, San Diego, La Jolla, CA, USA}

\author{Yongjun Chen}
\affiliation{%
  \institution{Salesforce Research}
\country{Palo Alto, CA, USA}
}

\email{yongjun.chen@salesforce.com}

\author{Zhiwei Liu}
\affiliation{%
  \institution{Salesforce Research}
\country{Palo Alto, CA, USA}
}
\email{zhiweiliu@salesforce.com}

\author{Jia Li}
\affiliation{%
  \institution{Salesforce Research}
  \country{Palo Alto, CA, USA}
  }

\email{jia.li@salesforce.com}

\author{Julian McAuley}
\affiliation{%
  \institution{UC San Diego}
  \country{La Jolla, CA, USA}
  }
\email{jmcauley@eng.ucsd.edu}

\author{Caiming Xiong}
\affiliation{%
  \institution{Salesforce Research}
  \country{Palo Alto, CA, USA}}
\email{cxiong@salesforce.com}

% \author{Anonymous authors}
%  \affiliation{
%   \institution{Paper under double-blind review}
%  }
% \renewcommand{\shortauthors}{Anonymous Author, et al.}

% \renewcommand{\shortauthors}{TODO, et al.}

\begin{abstract}

Users’ 
%behaviors of interacting 
interactions
with
%different 
items are
%often
driven by 
various intents 
(e.g., preparing for holiday gifts,
shopping for fishing equipment, etc.).
% Modeling such dependency,
% Thus, it is crucial 
% to consider users' intents 
% for providing a better recommendation.
However, users’ underlying intents 
are often unobserved/latent,
making it
%thus is 
%extremely
challenging 
to leverage such latent intents 
%and is less explored 
%in 
for
\emph{Sequential recommendation} (SR).
To 
%discover 
investigate
the benefits of latent intents
%factor,
and leverage them effectively for recommendation, 
we propose %namely 
\emph{\textbf{I}ntent \textbf{C}ontrastive  \textbf{L}earning} (\lpmname),
a general learning paradigm that
leverages a latent intent variable into SR.
%with use of user behavior sequences only.
The core idea is to learn users' intent distribution functions from unlabeled user behavior sequences
%themselves,
and optimize SR models with
%a new 
contrastive self-supervised learning 
(SSL) by considering the learnt intents
to 
improve recommendation.
Specifically, 
we introduce a
latent variable to represent users' intents
and learn the distribution function
of the latent variable via clustering.
We propose to 
leverage the learnt intents into SR models
%via a new 
via
contrastive SSL,
which maximizes the agreement
between a view of sequence 
and its corresponding intent.
The training is alternated between intent representation
learning and the SR model optimization steps
within the generalized expectation-maximization (EM) framework.
%, 
%which ensures convergence.
Fusing user intent information
into SR
also 
%benefits
%of improving 
improves
model  robustness.
%Theoretical analyses show the
%superiority of the proposed method.
Experiments conducted on four real-world datasets demonstrate the superiority of the proposed learning paradigm,
which improves performance,
and robustness
against data sparsity 
and noisy interaction issues
% Case studies on Sports and Yelp
% further verify the effectiveness of \lpmname
\footnote{Code is available at
\href{https://github.com/salesforce/ICLRec}{https://github.com/salesforce/ICLRec}}.
% even when recommender systems
% face
% heavy data-sparsity issues.

% Firstly,
% % The core idea is to leverage mutual intent 
% % information that is shared across users via contrastive SSL.
% we introduce intent prototypes as latent variables to represent users' intents 
% underlying their behavior sequences.
% Then we learn the intent representations via clustering,
% which are used to measure the correlation of sequences.
% Instead of
% maximizing the agreements of 
% a group of positive views directly,
% and maximize the agreement
% between a view and its corresponding intent 
% to improve robustness and computation efficiency.
% We alternate the above steps
% within the generalized expectation-maximization (EM) framework, which ensures convergence.
% Experiments conducted on four real-world datasets
% demonstrate the effectiveness of the proposed learning paradigm,
% which improves performance and robustness, 
% even when recommender systems
% %are facing 
% face
% heavy data-sparsity issues.
% users' interactions follow
%the number of user's behaviors follows 
% long-tail distributions. 

\end{abstract}



% \ccsdesc[500]{Information systems~Recommender systems}
% \ccsdesc[500]{Computing methodologies~Neural networks}




\maketitle

\section{Introduction}
Recommender systems have been widely 
used in many scenarios 
%(e.g., e-commerce, etc.) 
to provide personalized
items to users over massive vocabularies of items.
The core of an effective recommender system
is to accurately predict users'
interests toward
items based on their historical interactions.
With the success of deep learning,
deep \emph{Sequential Recommendation} (SR)~\cite{kang2018self,sun2019bert4rec} models, 
which aims at
dynamically characterizing
the behaviors of users
with different deep neural networks~\cite{yan2019cosrec,wu2017recurrent},
arguably represents the current state-of-the-art~\cite{kang2018self,sun2019bert4rec,ma2020disentangled,zhou2020s3,xie2020contrastive,li2020time,fan2021continuous}.

% Recent advances in deep learning motivate the developments of deep SR~\cite{kang2018self,sun2019bert4rec,zhou2020s3,xie2020contrastive} models,
% % Various models
% such as CNN-based~\cite{tang2018personalized,yan2019cosrec},
% RNN-based~\cite{hidasi2015session,wu2017recurrent}, 
% and Transformer-based ~\cite{kang2018self,li2020time} SR models. 

\begin{figure}[htb]
  \centering
  \includegraphics[width=0.9\linewidth]{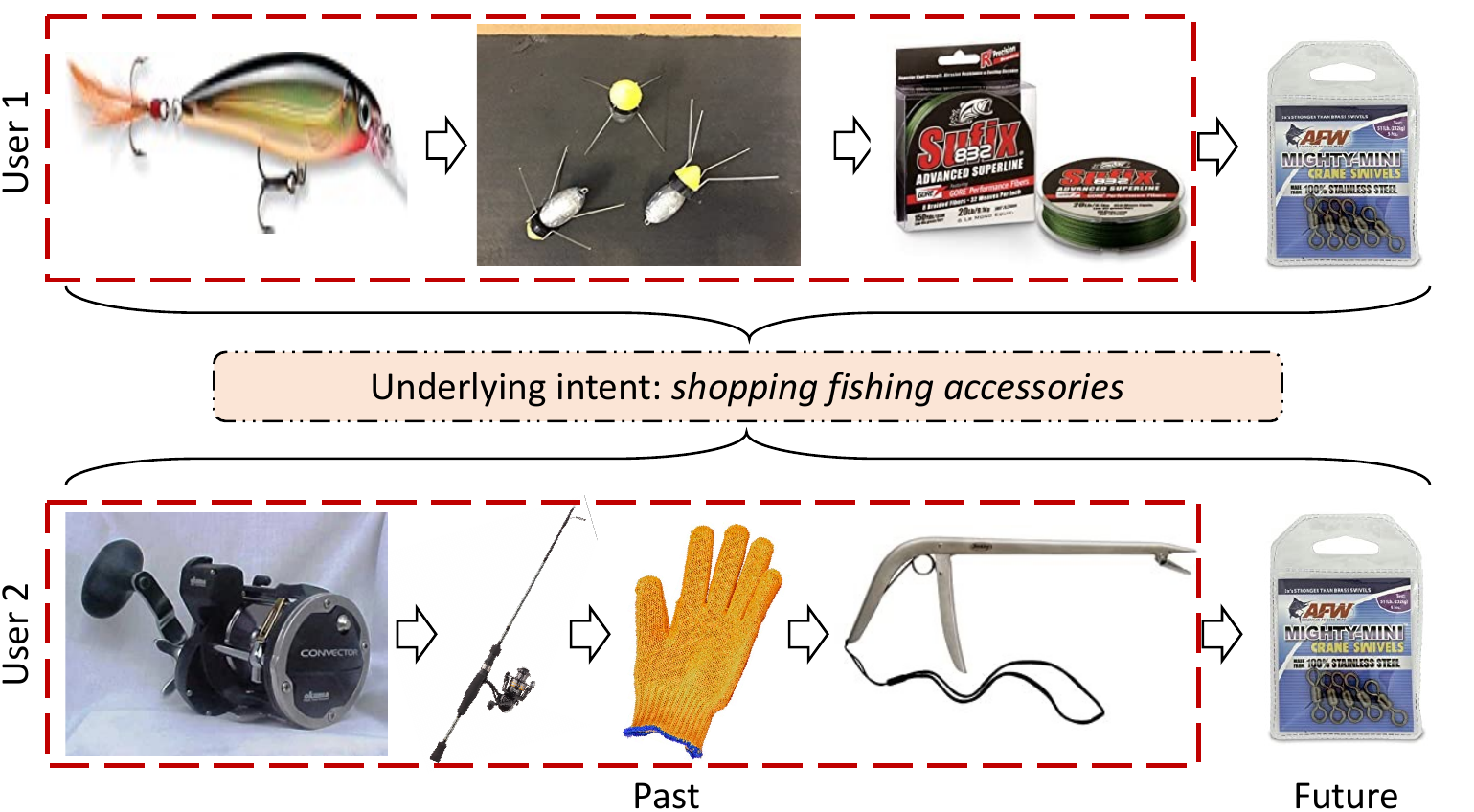}
  \caption{Users' purchasing behaviors 
  %on Amazon 
  can be driven by underlying intents that are not observed. }
  \label{fig:motivation}
\end{figure}

In general, a deep SR model is trained based on users' interaction
behaviors via a deep neural network,
% holding the believe that users' interests only
% depending on their historical behaviors. 
assuming users' interests depending on historical behaviors.
However, 
consuming behaviour of users can 
be affected by other latent factors,
i.e., driven by their underlying intents.
Consider the example illustrated in
Figure~\ref{fig:motivation}.
Two users purchased a series of different items on Amazon
in the past. 
Given such distinct interaction behaviors, the
system will recommend different items to them.
However, 
both of them are fishing enthusiasts
and are shopping
for fishing activities.
As a result,
they both purchase `fishing swivels'
in the future.
If the system is aware that these two users
are shopping
for fishing activities, then commonly purchased
items for fishing, such as 
`fishing swivels' can be suggested.
This motivates us to
mine underlying intents that are shared 
across users and use the learnt
intents to guide system providing recommendations.

Precisely discovering the intents of users, however, is under-explored. Most existing works~\cite{tanjim2020attentive,cai2021category}
of user intent modeling require side
information. 
ASLI~\cite{tanjim2020attentive} 
leverages user action types (e.g., click, add-to-favorite,
etc.) to capture users' intentions, whereas such information is not always available 
in system.
CoCoRec~\cite{cai2021category}
utilizes item category information.
But we argue that categorical feature is unable to accurately represents
users' intents.
For example, intents like
`shopping for holiday gifts' may involve 
items from multiple different categories.
DSSRec~\cite{ma2020disentangled} 
proposes a seq2seq training
strategy, which optimizes the intents 
in latent spaces.
However, those intents in DSSRec are inferred solely based on individual sequence representation, 
while ignoring the underlying correlations of the intents from different users. 

Effectively modeling latent intents 
from user behaviors
poses two challenges.
First, 
it is extremely difficult to learn latent intents accurately
because we have no labelling data for intents.
The only available supervision signals for intents are the user behavior data. Nevertheless, as aforementioned example indicates, distinct behaviors may reflect the same intent.
% Simply adding a deep neural network component
% on top of an SR model
% to optimize model parameters simultaneously 
% with estimating users' intents
% may lead to trivial solutions,
% in which all users' intents are inferred as one 
% type of intent.
Besides,
effectively fusing intent information
into a SR model is non-trivial.
The target in SR is to predict next items in sequences, which is solved by encoding sequences. 
Leveraging latent intents of sequences into the model requires the intent
factors to be orthogonal to the sequence embeddings, 
which otherwise would induce redundant information.

To discover the benefits of 
latent intents
and address challenges,
we propose the
\emph{\textbf{I}ntent \textbf{C}ontrastive \textbf{L}earning} (\lpmname),
a general learning paradigm that
leverages the latent intent 
factor into SR.
%with use of user behavior sequences 
%only.
It learns users' intent
distributions
from all user behavior sequences
via clustering.
And it leverages
the learnt intents
into the SR model 
via a new contrastive SSL,
which 
maximizes the agreement
between a view of sequence 
and its corresponding intent.
The intent representation learning module 
and the contrastive SSL module are mutually reinforced 
to train a more expressive
sequence encoder. 
We tackle the challenge of intent
mining problem by 
introducing a
latent variable to represent users' intents
and learn them alternately
along with the SR model optimization through
an expectation-maximization (EM) framework
to ensure convergence.
We suggest fusing learnt intent information
into SR via the proposed contrastive SSL,
as it can improve model's performance as well
as robustness.
% The key idea is to learn users' intent 
% distribution function 
% from unlabeled user behavior sequences 
% and optimize the SR model 
% via a new contrastive SSL 
% that considers the learned intents 
% to improve recommendation.
% Specifically, 
% we introduce a
% latent variable to represent users' intents
% and estimate the distribution function
% of the latent variable via clustering.
% We leverage the estimated intents into the SR model
% by proposing a new contrastive SSL objective,
% which 
% maximizes the agreement
% between a view of sequence 
% and its corresponding intent.
% We alternate the above steps
% within the generalized expectation-maximization (EM) framework, 
% which addresses the posed challenges and ensures convergence.
% Fusing user intent information
% into SR
% also improves 
% %benefits of improving 
% model robustness.
% We provide detailed theoretical analyses to show the
% superiority of the proposed method.
Extensive experiments conducted on four real-world datasets
further verify the effectiveness of the proposed learning paradigm,
which improves performance and robustness, 
even when recommender systems
face
heavy data sparsity issues. 

% The contributions of this work are summarized as follows:
% \begin{itemize}
%     \item We propose a novel \lpmname learning paradigm that
%     captures latent intents from user interaction behaviors,
%     and fuses them into an SR model via a new contrastive SSL objective.
%     \item We formulate \lpmname within the generalized EM framework
%     to guarantee convergence and provide detailed theoretical analyses.
%     \item We conduct extensive experiments on four datasets and case studies on two datasets,
%     which further verify the superiority of \lpmname.
% \end{itemize}

\section{Related Work}

% \subsection{Colleborative Filtering}
% TODO.

\subsection{Sequential Recommendation}

Sequential recommendation aims
to accurately characterize users' dynamic interests
by modeling their past behavior sequences~\cite{rendle2010factorization,rendle2010factorizing,kang2018self,chen2021modeling,li2021lightweight,liu2021contrastive}.
Early works on SR usually model 
an
item-to-item transaction pattern
based on Markov
Chains~\cite{rendle2010factorization,he2016fusing}.
FPMC~\cite{rendle2010factorizing} 
combines the advantages of Markov Chains 
and matrix factorization 
to fuse both sequential patterns 
and users' general interest.
With the recent advances of deep learning,
many deep sequential recommendation models
are also developed~\cite{tang2018personalized,hidasi2015session,kang2018self,sun2019bert4rec}. Such as Convolutional Neural Networks (CNN)-based~\cite{tang2018personalized} and 
RNN-based~\cite{hidasi2015session} models.
The recent success of Transformer~\cite{vaswani2017attention} 
also motivates
the developments of pure Transformer-based SR models.
SASRec~\cite{kang2018self} 
utilizes unidirectional Transformer 
to assign weights to each interacted item adaptively.
BERT4Rec~\cite{sun2019bert4rec} improves that by
utilizing a bidirectional Transformer
with a \emph{Cloze} task~\cite{taylor1953cloze}
to fuse user behaviors information from
left and right directions into each item.
LSAN~\cite{li2021lightweight} improves SASRec
on reducing model size perspective. 
It proposes a temporal context-aware embedding 
and twin-attention network, which are light weighted.
ASReP~\cite{liu2021augmenting} further alleviates
the data-sparsity issue 
by leveraging a pre-trained Transformer
on the revised user behavior sequences to 
augment short sequences. 
In this paper, we study the 
potential of addressing data sparsity issues 
and improving SR via self-supervised learning.

% \vspace{-0.5cm}
\subsection{User Intent for Recommendation}
Recently, many approaches have been proposed to study 
users' intents for 
improving recommendations~\cite{wang2019modeling,cen2020controllable,li2019multi,li2021intention}.
MCPRN~\cite{wang2019modeling}
designs mixture-channel purpose
routing networks to adaptively
learn 
users' different purchase purposes 
of each item
% purpose of each item
under different channels (sub-sequences) for session-based recommendation.
MITGNN\cite{liu2020basket}
proposes a 
multi-intent
translation graph neural network
to mine users' multiple intents
by considering the correlations of the intents.
ICM-SR~\cite{pan2020intent}
designs an
intent-guided neighbor detector
to retrieve correct
neighbor sessions
for neighbor representation.
Different from session-based recommendation,
another line of works
focus on modeling the sequential 
dynamics of users' interaction behaviors
in a longer time span.
% Different from session-based recommendation,
% users' interaction behaviors 
% in SR 
% often 
% %have under 
% cover
% a longer time span
% thus, users' intents are more complex.
DSSRec~\cite{ma2020disentangled} 
proposes a seq2seq training
strategy using multiple future interactions as supervision and introducing an intent variable from her historical and future behavior sequences. 
The intent variable is used to capture mutual information between an individual user's historical and future behavior sequences.
Two users of similar intents 
might be far away in representation space. 
% Users with similar intents are not necessarily
% to be close to each other in representation space.
% inferring high confidence sequences
% in the batch for seq2seq training, i.e., 
% a user that has consistent intent
% will be selected.
Unlike this work, our intent variable is learned over all users' sequences and is used to maximize mutual information across different users with similar learned intents.
ASLI~\cite{tanjim2020attentive}
captures intent
via a temporal convolutional
network with side information (e.g., user action types such as
click, add-to-favorite, etc.),
and then use the learned
intents to 
guide SR model to predict 
the next item. 
Instead, our method
can learn users' intents
based on user interaction 
%behaviors
data only.

\subsection{Contrastive Self-Supervised Learning}
\label{sec:contrastive-ssl}
% \hspace{1pt}

Contrastive Self-Supervised Learning (SSL)
has brought much attentions by
different research communities
including CV~\cite{chen2020simple,li2020prototypical,he2020momentum,caron2020unsupervised,khosla2020supervised} and
NLP~\cite{gao2021simcse,gunel2020supervised,mnih2013learning,zhang2020unsupervised},
%And it also starts to influence
as well as
recommendation
%community
%recently.
\cite{yao2020self,zhou2020s3,wu2021self,xie2020contrastive}.
The fundamental goal of contrastive SSL
is to maximize mutual information
among the positive transformations 
of the data itself while
improving
%model's 
discrimination ability
to the negatives.
In reccommendation,
A two-tower DNN-based
contrastive SSL 
model is proposed in~\cite{yao2020self}.
It
aims
to
improving collaborative filtering 
based
recommendation leveraging item attributes.
SGL~\cite{wu2021self} adopts
a multi-task framework with 
contrastive SSL to improve the
graph neural networks (GCN)-based
collaborative filtering methods~\cite{he2020lightgcn,wang2019neural,liu2020deoscillated,zhang2020stacked}
with only item IDs as features.
Specific to SR,
% Seq2Seq~\cite{ma2020disentangled};
% S$^3\text{-Rec}$~\cite{zhou2020s3};
S$^3\text{-Rec}$~\cite{zhou2020s3}
adopts a pre-training and 
fine-tuning strategy, and utilizes
contrastive SSL during pre-training
to incorporate correlations
among items, sub-sequences, and attributes of a given
user behavior sequence. 
However, the two-stage training strategy
prevents the information sharing between next-item prediction and SSL tasks and restricts
the performance improvement.
CL4SRec~\cite{xie2020contrastive} and 
CoSeRec~\cite{liu2021contrastive}
instead
utilize a multi-task training framework 
%instead
with a contrastive objective 
to enhance user representations.
Different from them, our work is aware of
users' latent intent factor when
leveraging contrastive SSL,
which we show to be
beneficial for 
improving recommendation 
performance 
and robustness.

\section{Preliminaries}
% \hspace{1pt}
\subsection{Problem definition}
\label{sec:problem-definition}

Assume that a recommender system has a set of users 
and items denoted by $\mathcal{U}$ and $\mathcal{V}$ respectively. 
% $|\mathcal{U}|$ and $|\mathcal{V}|$ are the numbers of users and items, respectively.
Each user $u \in \mathcal{U}$ has a sequence of interacted items
sorted in chronological order $S^{u}= [s^{u}_{1}, \dots, s^{u}_{t}, \dots, s^{u}_{|S^{u}|}]$
where $|S^{u}|$ is the number of interacted items
and $s^{u}_{t}$ is the item $u$ interacted at
step $t$. We denote $\mathbf{S}^{u}$ 
as embedded representation of $S^{u}$,
where $\mathbf{s}^{u}_{t}$ is the d-dimensional embedding of item $s^{u}_{t}$. 
In practice, sequences are truncated with
maximum length $T$.
If the sequence length is greater than $T$, the most
recent $T$ actions are considered. If the sequence length is less than $T$, `padding' items will be 
added to the left until the length is
$T$~\cite{tang2018personalized,hidasi2015session,kang2018self}. 
For each user $u$, 
the goal of next item prediction task is to predict
the next item that the user $u$ 
is most likely to interact with 
at the $|S_{u}|+1$ step among the item set $\mathcal{V}$,
given sequence $\mathbf{S}^{u}$.
% which is formulated as follows:
% \begin{equation}
% \underset{v_{i}\in \mathcal{V}}{\mathrm{arg\,max}}~P(v_{|s_{u}|+1}=v_{i}\left| s_{u}\right.),
% \end{equation}
\hspace{1pt}
\subsection{Deep SR Models for Next Item Prediction}

Modern sequential recommendation models 
commonly encode user behavior sequences with a deep neural network
to model sequential patterns
from (truncated)
user historical behavior sequences.
Without losing generality, we define a sequence encoder $f_{\theta}(\cdot)$ that
encodes a sequence $\mathbf{S}^{u}$ 
and outputs user interest representations over all 
position steps $\mathbf{H}^{u}=f_{\theta}(\mathbf{S}^{u})$. 
Specially, $\mathbf{h}^{u}_{t}$ represents user's interest at position $t$. 
% In this work, we use Transformer~\cite{vaswani2017attention} architecture to encode sequences 
% as it has been successful in SR~\cite{kang2018self,sun2019bert4rec,zhou2020s3,ma2020disentangled}. 
The goal can be formulated as finding the optimal 
encoder parameter $\theta$ that maximizes the log-likelihood
function of the expected next items of given $N$ sequences on all positional steps:
\begin{equation}
\label{eq:problem_define}
\theta^{*} = \underset{\theta}{\mathrm{arg\,max}}~\sum_{u=1}^{N} \sum_{t=2}^{T}
\ln P_{\theta}(s^{u}_{t}).
\end{equation}

% We can see that maximizing Eq.~\ref{eq:problem_define}
% is equivalent to minimizing following 
% objective:
% \begin{equation}
% \label{eq:problem_define2}
% \theta^{*} = \underset{\theta}{\mathrm{arg\,min}} -\sum_{u=1}^{N} \sum_{t=2}^{T}
% \ln P_{\theta}(s^{u}_{t}),
% \end{equation}

which is equivalent to minimizing the adapted binary cross-entropy loss as follows:
\begin{equation}
\label{eq:next-item}
\mathcal{L}_{\mathrm{NextItem}} = \sum_{u=1}^{N} \sum_{t=2}^{T} \mathcal{L}_{\mathrm{NextItem}}(u,t),
\end{equation}

\begin{equation}
\label{eq:next-item-2}
\mathcal{L}_{\mathrm{NextItem}}(u,t) = -\log(\sigma(\mathbf{h}^{u}_{t-1}\cdot \mathbf{s}^{u}_{t}))- \sum_{neg}\log(1-\sigma (\mathbf{h}^{u}_{t-1} \cdot \mathbf{s}^{u}_{neg})),
\end{equation}
where $\mathbf{s}^{u}_{t}$ and $\mathbf{s}^{u}_{neg}$ 
denote the embeddings of the target item $s_{t}$ and all
items
not interacted by $u$.
The sum operator in Eq.~\ref{eq:next-item-2} is 
computationally expensive because $|V|$ is large.
Thus we follow~\cite{kang2018self,zhou2020s3,cen2020controllable} to use a sampled softmax technique to randomly sample a negative item for each time step in each sequence.
% randomly sampled negative item(s) from $V$ excluding items that $u$
% has interacted with, respectively. 
$\sigma$ is the sigmoid function. 
And $N$ is refers to the
mini-batch size as the SR model.

\subsection{Contrastive SSL in SR}

\begin{figure*}[htb]
  \centering
  \includegraphics[width=0.95\textwidth]{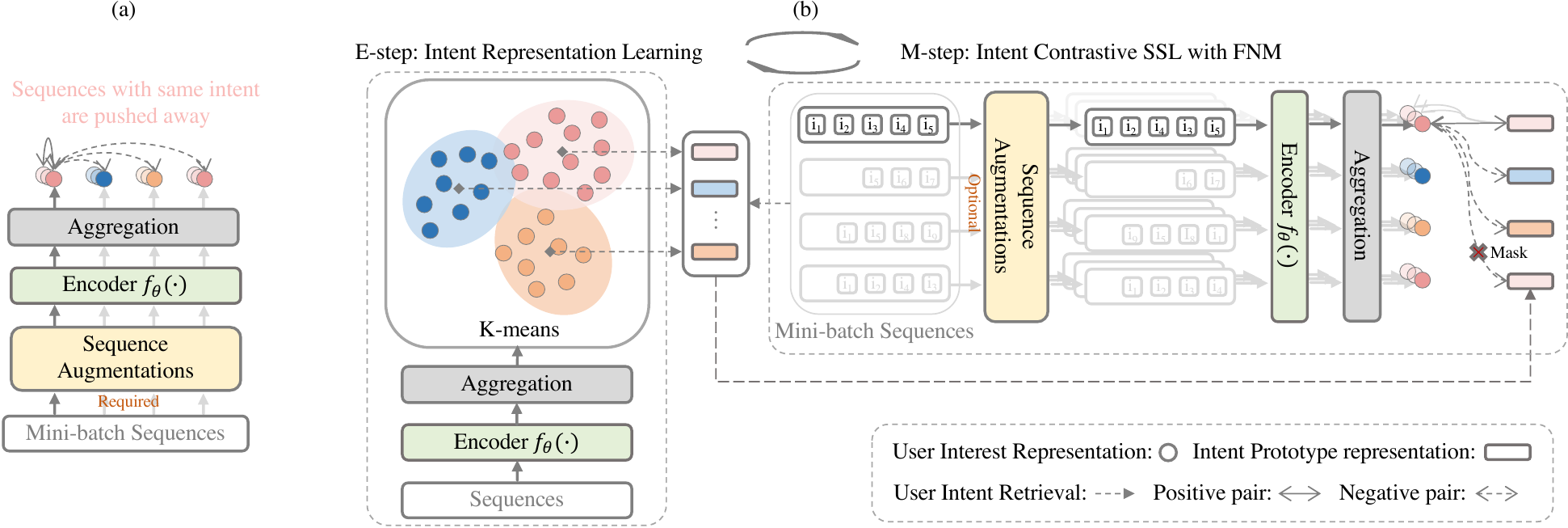}
  \caption{Overview of \lpmname.  (a) An individual sequence level SSL for SR. 
  (b) The
  proposed \lpmname for SR. 
  It alternately performs intent representation
  learning and intent contrastive SSL with FNM within the generalized EM framework to maximizes mutual
  information (MIM) 
  between a behavior sequence and its
  corresponding intent prototype.
  }
  \label{fig:ssl_comparison}
\end{figure*}

Recent advances in contrastive SSL
%inspires 
have inspired the
recommendation community
to leverage contrastive SSL to
fuse correlations among 
different views of one sequence~\cite{chen2020simple,yao2020self,wu2021self},
following the
mutual information maximization (MIM) principle.
Existing approaches in SR
can be seen as
instance discrimination tasks
that optimize a lower bound of MIM,
such as InfoNCE~\cite{oord2018representation,he2020momentum,chen2020simple,li2020prototypical}.
It aims to optimize the
proportion of gap of positive pairs and negative pairs~\cite{liu2021self}.
In such an instance discrimination task,
sequence augmentations such as `mask', `crop', or `reorder' are required to
create 
different views of the unlabeled data in SR~\cite{sun2019bert4rec,zhou2020s3,xie2020contrastive,zhou2021pure}.
Formally, given a sequence $S^{u}$,
and a pre-defined data transformation 
function set $\mathcal{G}$, we can create
two positive views of $S^{u}$ as follows:
\begin{equation}
\label{eq:data-augmentation}
\Tilde{S}^{u}_{1} = g_{1}^{u}(S^{u}), \Tilde{S}^{u}_{2} = g_{2}^{u}(S^{u}), \text{ s.t. }g_{1}^{u}, g_{2}^{u} \sim \mathcal{G},
\end{equation} 
where $g_{1}^{u}$ and $g_{2}^{u}$ are transformation functions sampled 
from $\mathcal{G}$ to create
a different view of sequence $s_{u}$. 
Commonly, views created from the same sequence
are treated as positive pairs, 
and the views of any different sequences
are considered as negative pairs. 
The augmented views are first encoded with the
sequence encoder $f_{\theta}(\cdot)$ to 
$\mathbf{\Tilde{H}}^{u}_{1}$ and $\mathbf{\Tilde{H}}^{u}_{2}$,
and then be fed into an `Aggregation'
layer to get vector representations
of sequences, denoted as $\mathbf{\Tilde{h}}^{u}_{1}$ and $\mathbf{\Tilde{h}}^{u}_{2}$. In this paper,
we `concatenate' users' interest representations over time steps
for simplicity. Note that sequences are prepossessed to have the same length (See Sec.~\ref{sec:problem-definition}), thus
their vector representations after concatenation 
have the same length too.
After that,
we can optimize $\theta$ via InfoNCE loss:
\begin{equation}
\label{eq:info-nce1}
\mathcal{L}_{\mathrm{SeqCL}} = \mathcal{L}_{\mathrm{SeqCL}}(\mathbf{\Tilde{h}}^{u}_{1}, \mathbf{\Tilde{h}}^{u}_{2}) +
\mathcal{L}_{\mathrm{SeqCL}}(\mathbf{\Tilde{h}}^{u}_{2}, \mathbf{\Tilde{h}}^{u}_{1}),
\end{equation}  
and 
\begin{equation}
\label{eq:info-nce2}
\mathcal{L}_{\mathrm{SeqCL}}(\mathbf{\Tilde{h}}^{u}_{1}, 
                    \mathbf{\Tilde{h}}^{u}_{2}) = 
        - \log \frac{\exp(\text{sim}(\mathbf{\Tilde{h}}^{u}_{1}, \mathbf{\Tilde{h}}^{u}_{2}))}
        {\sum_{neg}\exp(\text{sim}(\mathbf{\Tilde{h}}^{u}_{1}, \mathbf{\Tilde{h}}_{neg}))},
\end{equation}  
where $sim(\cdot)$ is dot product and 
$\mathbf{\Tilde{h}}_{neg}$ are negative
views' representations of sequence $S^{u}$.
Figure~\ref{fig:ssl_comparison}
(a) illustrates how SeqCL  works.

\subsection{Latent Factor Modeling in SR}
\label{subsec:lfm}
The main goal of next item prediction task is to optimize
Eq.~(\ref{eq:problem_define}).
% , assuming that
% users' next interactions with items only depends
% on users' explicit behaviors $S^{u}$. 
% However, there might have some latent factors
% that affect
% users' decision-making when interacting with next items.
% Thus, it is important to leverage users' intents
% to provide better recommendations. 
Assume that there are also $K$ different user intents (e.g., purchasing holiday gifts, preparing for fishing activity, etc.)
in a recommender system that forms 
the intent variable $c=\left\{ c_{i} \right\}_{i=1}^{K}$, then
the probability of a user interacting with a certain
item can be rewritten as follows:
\begin{equation}
\label{eq:3}
\begin{split}
P_{\theta}(s^{u})= 
\mathbb{E}_{(c)} \left[
P_{\theta}(s^{u}, c)
\right].
\end{split}
\end{equation}
% However, users' intents are latent by definition.
% As we can observe users' behaviors
% but the underlying intents are often unobserved.
However, users intents are latent by definition.
% As we can observe users' behaviors
% but the underlying intents are often unobserved.
Because of the missing observation of variable $c$,
we are in a `chicken-and-eggs' situation that
without $c$, we cannot estimate parameter $\theta$,
and without $\theta$ we cannot infer
what the value of $c$ might be.

%Fortunately, 
Later, we will show that a generalized Expectation-Maximization framework
provides a direction to address above problem
with a convergence guarantee. 
The basic idea of optimizing Eq.~(\ref{eq:3}) via EM 
is to start with an initial guess 
of the model parameter $\theta$
and estimate the expected values
of the missing variable $c$, i.e., the E-step.
And once we have the values of $c$,
we can maximize the Eq.~(\ref{eq:3}) w.r.t the
parameter $\theta$, i.e., the M step.
We can repeat this iterative process until the likelihood cannot increase anymore.

\section{Method}

The overview of the proposed \lpmname within 
EM framework is presented in Figure~\ref{fig:ssl_comparison} (b). 
It performs E-step and M-step alternately
to estimate the distribution function $Q(c)$ over the intent variable $c$ and optimize the model parameters
$\theta$. 
In E-step, it 
estimates $Q(c)$ via clustering.
In M-step, it optimizes $\theta$ with considering
the estimated $Q(c)$ via mini-batch gradient descent.
In each iteration, $Q(c)$ and 
$\theta$ are updated.

In the following sections, 
we first derive the objective function in order
to model
the latent intent variable $c$ into an SR model, 
and how to alternately
optimize the objective function w.r.t. $\theta$
and estimate the distribution of $c$
under 
a generalized EM framework in Section~\ref{subsec:ica}.
Then we describe the overall training
strategy in Section~\ref{subsec:optimization}.
We provide detailed analyses in Section~\ref{subsec:discussion}
followed by experimental studies in Section~\ref{sec:experiments}.

% In this section,
% we first introduce a novel 
% learning paradigm \lpmname,
% which aims
% at maximize mutual information
% across sequences via contrastive SSL.
% And then explain the 

% \begin{figure*}[htb]
%   \centering
%   \includegraphics[width=0.9\textwidth]{./figures/architecture.pdf}
%   \caption{The Overview of ICLRec framework. (TO BE DONE. Maybe consider draw S3Rec, Seq2Seq, and CL4SRec figure for comparison, refer BERT4Rec figure.) )}
%   \label{fig:overall-architecture}
% \end{figure*}

\subsection{Intent Contrastive Learning}
\label{subsec:ica}
\subsubsection{\textbf{Modeling Latent Intent for SR}}
Assuming that there are $K$
latent intent prototypes $\left\{ c_{i} \right\}_{i=1}^{K}$ that affect users' decisions to
interact with items,
then based on Eq.~(\ref{eq:problem_define}) and~(\ref{eq:3}), we can rewrite objective as follows:
\begin{equation}
\label{eq:mle0}
\begin{split}
\theta^{*} = \underset{\theta}{\mathrm{arg\,max}} \sum_{u=1}^{N} \sum_{t=1}^{T}
\ln \mathbb{E}_{(c)} \left[
P_{\theta}(s^{u}_{t}, c_{i})
\right],
\end{split}
\end{equation}
which is however hard
to optimize.
Instead, we construct a lower-bound function
of Eq.~(\ref{eq:mle0}) and maximize the lower-bound.
Formally,
assume intent $c$ follows distribution $Q(c)$, where
$\sum_{c} Q(c_{i})=1$ and $Q(c_{i})\geq 0$. 
Then we have
\begin{equation}
\label{eq:mle1}
\begin{split}
\sum_{u=1}^{N} \sum_{t=1}^{T}
\ln \mathbb{E}_{(c)} \left[
P_{\theta}(s^{u}_{t}, c_{i}) 
\right] =\sum_{u=1}^{N} \sum_{t=1}^{T}
\ln \sum_{i=1}^{K}
P_{\theta}(s^{u}_{t}, c_{i}) 
 \\
= \sum_{u=1}^{N} \sum_{t=1}^{T}
\ln \sum_{i=1}^{K}
Q(c_{i})\frac{P_{\theta}(s^{u}_{t}, c_{i})}{Q(c_{i})}.
\end{split}
\end{equation}
Based on the Jensen’s inequality, the term in Eq.~(\ref{eq:mle1}) is

\begin{equation}
\label{eq:mle1-2}
\begin{split}
\geq \sum_{u=1}^{N} \sum_{t=1}^{T}
\sum_{i=1}^{K} Q(c_{i})\ln
\frac{P_{\theta}(s^{u}_{t}, c_{i})}{Q(c_{i})} \\
\propto \sum_{u=1}^{N} \sum_{t=1}^{T} \sum_{i=1}^{K} Q(c_{i}) \cdot \ln P_{\theta}(s^{u}_{t}, c_{i}),
\end{split}
\end{equation}
where the $\propto$ stands for `proportional to' (i.e. up to a multiplicative constant). 
The inequality will hold with equality when
$Q(c_{i})=P_{\theta}(c_{i}| s^{u}_{t})$. 
% (Appendix~\ref{sec:lower-bound} provides full steps of derivation.) 
For simplicity, we only focus on last positional step 
when optimize
the lower-bound, which is defined as:
\begin{equation}
\label{eq:mle4}
\begin{split}
\sum_{u=1}^{N} \sum_{i=1}^{K} Q(c_{i}) \cdot \ln P_{\theta}(S^{u}, c_{i}),
\end{split}
\end{equation}
where $Q(c_{i})=P_{\theta}(c_{i}|S^{u})$. 

So far, we have found a lower-bound of Eq.~(\ref{eq:mle0}). However,
we cannot directly optimize Eq.~(\ref{eq:mle4})
because $Q(c)$ is unknown. 
Instead, we alternately optimize the model between 
the Intent Representation Learning (E-step)
and the Intent Contrastive SSL with FNM (M-step), which follows a generalized EM framework.
We term the whole processes Intent Contrastive 
Learning (\lpmname).
In each iteration, $Q(c)$ and the model parameter
$\theta$ are updated.

\subsubsection{\textbf{Intent Representation Learning}}
\label{sec:intent-represent-learning}

To learn the intent distribution function $Q(c)$, 
we encode all the
sequences $\left\{ S^{u}\right\}_{u=1}^{|\mathcal{U}|}$ with the encoder ${\theta}$ followed by
an `aggregation layer',
and then we perform $K$-means clustering over all 
sequence representations $\left\{\mathbf{h}^{u} \right\}^{|\mathcal{U}|}_{u=1}$ to obtain $K$ clusters.
After that, 
we can define the distribution function $Q(c_{i})$
as follows:
\begin{equation}
\label{eq:q-distribution}
Q(c_{i}) = P_{\theta}(c_{i}|S^{u}) = \left\{\begin{matrix}
1& \text{if $S^{u}$ in cluster $i$}  \\ 
0& \text{else}.
\end{matrix}\right.
\end{equation}
We denote $\mathbf{c_{i}}$ as the vector 
representation of intent $c_{i}$, which 
is the centroid representation of the $i^{th}$ cluster.
In this paper, we use `aggregation layer' to denote the the mean pooling operation
over all position steps for simplicity.
We leave other advanced aggregation methods such as attention-based methods
for future work studies.
Figure~\ref{fig:ssl_comparison} (b) illustrates 
how the E-step works.

\subsubsection{\textbf{Intent Contrastive SSL with FNM}}
\label{secintent-contrast}
% Quickly go through augmentation to create positive views of the original sequences. and mention, in general we can create multiple positive views to benefits training.

% On the other hand,two distinct sequence with diverse items may also be correlated ifthey share similar underlying intentions. 

% how to define positive pairs: K controls

% a positive group where: contrastive between 

We have estimated the distribution function
$Q(c)$. 
To maximize Eq.~(\ref{eq:mle4}), we also need
to define $P_{\theta}(S^{u}, c_{i})$.
Assuming that the prior over intents follow the uniform distribution and 
the conditional distribution
of $S^{u}$ given $c$ is isotropic Gaussian
with $L2$ normalization,
then we can rewrite $P_{\theta}(S^{u}, c_{i})$ as follows:
\begin{equation}
\label{eq:prior-probability}
\begin{split}
P_{\theta}(S^{u}, c_{i}) & = P_{\theta}(c_{i})P_{\theta}(S^{u}|c_{i}) =  \frac{1}{K} \cdot P_{\theta}(S^{u}| c_{i}) \\
& \propto \frac{1}{K} \cdot 
  \frac{\exp(-(\mathbf{h}^{u}-\mathbf{c}_{i})^2)}{\sum_{j=1}^{K}\exp(-(\mathbf{h}^{u}_{i}-\mathbf{c}_{j})^{2})} \\
& \propto \frac{1}{K} \cdot 
      \frac{\exp(\mathbf{h}^{u}\cdot \mathbf{c}_{i})}{\sum_{j=1}^{K}\exp(\mathbf{h}^{u}\cdot \mathbf{c}_{j})}, \\
\end{split}
\end{equation}
where $\mathbf{h}^{u}$ and $\mathbf{c}_{u}$ are vector representations
of $S^{u}$ and $c_{i}$, respectively. Based on Eq.~(\ref{eq:mle4}),~(\ref{eq:q-distribution}),~(\ref{eq:prior-probability}), maximizing Eq.~(\ref{eq:mle4}) is equivalent to minimize the following loss function:

\begin{equation}
\label{eq:icl-loss-basic} -\sum_{v=1}^{N}\log \frac{\exp(\text{sim}(\mathbf{h}^{u}, \mathbf{c}_{i}))}{\sum_{j=1}^{K}\exp(\text{sim}(\mathbf{h}^{u}, \mathbf{c}_{j}))},
\end{equation}
where $\text{sim}(\cdot)$ is a dot product.
We can see that Eq.~(\ref{eq:icl-loss-basic}) has 
a similar form as Eq.~(\ref{eq:info-nce2}),
where Eq.~(\ref{eq:info-nce2}) tries to maximize
mutual information between two individual sequences.
While Eq.~(\ref{eq:icl-loss-basic})
maximizes mutual information between
one individual sequence and its corresponding
intent. Note that,
sequence augmentations are required in SeqCL
to create positive views for Eq.~(\ref{eq:info-nce2}).
While in \lpmname, sequence augmentations are optional,
as the view of a given sequence is its corresponding 
intent that learnt from original dataset.
% Ideally, optimizing 
% Eq.~(\ref{eq:icl-loss-basic}) w.r.t. $\theta$
% will not require data augmentations if the training set is large enough. The data augmentations in
% existing contrastive SSL are
% used to create 
% positive views while in \lpmname,
% positive views can be identified 
% from the original dataset
% based on the learnt intents.
In this paper,
we apply sequence augmentations
for enlarging training set purpose and optimize model w.r.t $\theta$ based on Eq.~(\ref{eq:icl-loss-basic}).
Formally, given a batch of training sequences $\left \{ s_{u} \right \}_{u=1}^{N}$, we first create two
positive views of a sequence via Eq.~(\ref{eq:data-augmentation}), and then 
optimize the following loss function:

\begin{equation}
\label{eq:icl-general}
\mathcal{L}_{\mathrm{\lpmname}} =
                    \mathcal{L}_{\mathrm{\lpmname}}(
                    \mathbf{\Tilde{h}}^{u}_{1}, 
                    \mathbf{c}_{u})
                    +
                    \mathcal{L}_{\mathrm{\lpmname}}(
                    \mathbf{\Tilde{h}}^{u}_{2}, 
                    \mathbf{c}_{u}),
\end{equation}

and 
\begin{equation}
\label{eq:icl-single}
        \mathcal{L}_{\mathrm{\lpmname}}(
        \mathbf{\Tilde{h}}^{u}_{1}, 
        \mathbf{c}_{u})= 
        - \log \frac{\exp(\text{sim}(\mathbf{\tilde{h}}^{u}_{1},                       \mathbf{c}_{u}))}
        {\sum_{neg}\exp(\text{sim}(\mathbf{\tilde{h}}^{u}_{1}, \mathbf{c}_{neg}))},
\end{equation} 
where $c_{neg}$ are all the intents in the given batch.
However, directly optimizing Eq.~(\ref{eq:icl-single})
can introduce false-negative samples
since users in a batch can have same 
intent.
To mitigate the effects of false-negatives,
we propose a simple strategy
to mitigate the effects by not contrasting against them:
\begin{equation}
\label{eq:icl-single-de-noise}
        \mathcal{L}_{\mathrm{\lpmname}}(
        \mathbf{\Tilde{h}}^{u}_{1}, 
        \mathbf{c}_{u})
        = 
        - \log \frac{\exp(\text{sim}(\mathbf{\tilde{h}}^{u}_{1},                       \mathbf{c}_{u}))}
        {\sum_{v=1}^{N}\mathbbm{1}_{v\notin \mathcal{F}}\exp(\text{sim}(\mathbf{\tilde{h}}_{1}, \mathbf{c}_{v}))},
\end{equation} 

where $\mathcal{F}$ is a set of users that have same intent as $u$ in the mini-batch.
We term this
\textbf{False-Negative Mitigation (FNM)}.
Figure~\ref{fig:ssl_comparison} (b) illustrates 
how the M-step works.

\subsection{Multi-Task Learning}
\label{subsec:optimization}

We train the SR model with a multi-task training strategy 
to jointly optimize \lpmname via Eq.~(\ref{eq:icl-single-de-noise}), the main 
next-item prediction task via Eq.~(\ref{eq:next-item}) and a sequence level SSL task via Eq.~(\ref{eq:info-nce1}).
Formally, we jointly train the SR model $f_{\theta}$ 
as follows:

\begin{equation}
\label{eq:multi-task}
\mathcal{L}= \mathcal{L}_{\text{NextItem}} + \lambda \cdot \mathcal{L}_{\text{\lpmname}}
                + \beta \cdot \mathcal{L}_{\text{SeqCL}},
\end{equation} 
where $\lambda$ and $\beta$ control the strengths of the \lpmname
task
and sequence level SSL tasks, respectively. Appendix~\ref{sec：algorithm} provides the pseudo-code
of the entire learning pipeline. Specially, we build the learning paradigm on
Transformer~\cite{vaswani2017attention,kang2018self} encoder to form the model \modelwithtansformer.

\lpmname  is a model-agnostic objective, 
so we also apply it to 
S$^3\text{-Rec}$~\cite{zhou2020s3} 
model, which is pre-trained with several $\mathcal{L}_{\text{SeqCL}}$
objectives to capture correlations among
items, associated attributes, and subsequences
in a sequence and fine-tuned with the $\mathcal{L}_{\text{NextItem}}$ objective,
to further verify its effectiveness (see~\ref{sec:ablation} for details). 

% It can also be applied to other SR models
% that encode sequences into continuous representation space.
% Such as Caser~\cite{tang2018personalized} and GRU4Rec~\cite{hidasi2015session}.

\subsection{Discussion}
\label{subsec:discussion}

\subsubsection{\textbf{Connections with Contrastive SSL in SR}}
Recent methods~\cite{zhou2020s3,xie2020contrastive} in SR
follow standard contrastive SSL
to maximize mutual
information between two positive 
views of sequences. 
For example, CL4SRec encodes sequences with
Transformer and maximizes
mutual information between two
sequences that are augmented 
(cropping, masking, or reordering) from the
original sequence.
However, if the item
relationships of a sequence 
are vulnerable to 
random perturbation,
two views of this sequence 
may not reveal the original
sequence correlations.
\modelwithtansformer maximizes
mutual information between a sequence
and its corresponding intent prototype.
Since the intent prototype can be considered
as a positive view of a given sequence that learnt by
considering
the semantic structures of all sequences,
which reflects true sequence correlations,
the \modelwithtansformer can outperform
CL4SRec consistently.

% Another line of works are
% on considering latent intent for SR.
% For example,
% ASLI~\cite{tanjim2020attentive} assumes that
% users' intents are 
% conditionally independent 
% to their interacted items
% and their action types (click, add-to-favorite, etc).
% while \lpmname learn users'
% latent intents based only on user
% interaction data.

% \vspace{-10pt}
\subsubsection{\textbf{Time Complexity and Convergence Analysis}}
In every iteration of the training phase, 
the computation costs 
of our proposed method
are mainly from
the E-step estimation of $Q(\cdot)$ and M-step optimization of $\theta$
with multi-tasks training.
For the E-step, 
the time complexity is $O(|U|mKd)$ from clustering, where $d$ is the dimensionality
of the embedding and $m$ is the
maximum iteration number in clustering ($m=20$ in this paper).
For the M-step,
since we have three objectives
to optimize the network $f_{\theta}(\cdot)$,
the time complexity is $O(3\cdot(|U|^{2}d+|U|d^{2})$.
The overall complexity is dominated by the term $O(3\cdot(|U|^{2}d))$,
which is 3 times of Transformer-based SR with only next item prediction objective, e.g., SASRec. 
Fortunately, 
the model can be effectively parallelized because
$f_{\theta}$ is Transformer and we leave it in future work.
In the testing phase,
the proposed \lpmname as well
as the SeqCL objectives
are no longer needed, which yields the
model to have the same
time complexity as SASRec ($O(d|V|)$).
The empirical time spending comparisons
are reported in Sec.~\ref{subsec:results}. 
The convergence of \lpmname is guaranteed
under the generalized EM framework.
Proof is provided in Appendix~\ref{sec:em-convergence}.

%\section{Experimental Studies}
\section{Experiments}
\label{sec:experiments}

\begin{table*}[htb]
  \caption{Performance comparisons of different methods. 
  %Bold score is the best 
  %in each row, and underlined score is the second best
  The best score is bolded in each row, and the second best is underlined.
%   results across all approaches. 
  The last two columns are the relative
  improvements compared with the best baseline results.}
  \label{tab:main-results}
  \setlength{\tabcolsep}{1.3mm}{
  \begin{tabular}{ll|c|ccc|c|ccc|c|r}
    \toprule
    \multirow{1}{*}{Dataset} & \multirow{1}{*}{Metric} & \multirow{1}{*}{BPR} & 
    \multirow{1}{*}{GRU4Rec} & \multirow{1}{*}{Caser} &
    \multirow{1}{*}{SASRec} &
    \multirow{1}{*}{DSSRec} & \multirow{1}{*}{BERT4Rec} &
    \multirow{1}{*}{S$^3\text{-Rec}_{ISP}$} & \multirow{1}{*}{CL4SRec} &
    \multirow{1}{*}{\modelwithtansformer} 
    & Improv. \\
    % & & & & & & & & & & & SASRec & CL4SRec \\
    
    % \multicolumn{2}{c|}{Hit Rate} & \multicolumn{2}{c|}{NDCG} &\multicolumn{1}{c|}{MAP}&\multicolumn{1}{c}{Avg.}  \\

    % Dataset & Metric & PopRec & BPR  & GRU4Rec & Caser & SASRec & BERT4Rec & S$^3$Rec-MIPSP & CL4SRec & CoSeRec & \multicolumn{2}{c}{Improv.} \\
    \hline
    % & & & & & & & & & & SASRec & All \\
    \midrule
    \multirow{4}{*}{Sports}  
    & HR@5    & 0.0141 & 0.0162 & 0.0154 & 0.0206 & 0.0214 & 0.0217 & 0.0121 & \underline{0.0217}$\pm$0.0021 & \textbf{0.0283}$\pm$0.0006 & 30.48\% \\
                            %  & HR@10   & 0.0094 & 0.0216 & 0.0258 & 0.0261 & 0.0320 & 0.0286 & 0.0205 & \underline{0.0369} & \textbf{0.0437} & 18.43\%\\
                             & HR@20   & 0.0323 & 0.0421 & 0.0399 & 0.0497& 0.0495 & \underline{0.0604} & 0.0344 & 0.0540$\pm$0.0024 &  \textbf{0.0638}$\pm$0.0023 & 18.15\% \\
                             & NDCG@5  & 0.0091 & 0.0103 & 0.0114 & 0.0135 & 0.0142 & 0.0143 & 0.0084 & \underline{0.0137}$\pm$0.0013 &     
                        \textbf{0.0182}$\pm$0.0001 & 33.33\%\\
                            %  & NDCG@10 & 0.0053 & 0.0115 & 0.0142 & 0.0135 & 0.0172 & 0.0141 & 0.0111 & \underline{0.0191} & \textbf{0.0234} & 22.51\% \\
                             & NDCG@20 & 0.0142 & 0.0186 & 0.0178 & 0.0216 & 0.0220 & \underline{0.0251} & 0.0146 & 0.0227$\pm$0.0016 &  \textbf{0.0284}$\pm$0.0008 & 24.89\% \\
    \midrule
    \multirow{4}{*}{Beauty}  
    & HR@5   & 0.0212  & 0.0111 & 0.0251 & 0.0374 & 0.0410 & 0.0360 & 0.0189 & \underline{0.0423}$\pm$0.0031  & \textbf{0.0493}$\pm$0.0013 & 16.43\% \\
                            %  & HR@10  & 0.0152 & 0.0372 & 0.0162 & 0.0342 & 0.0575 & 0.0601 & 0.0307 & \underline{0.0694} & \textbf{0.0737} & 6.20\%\\
                             & HR@20  & 0.0589 & 0.0478 & 0.0643 & 0.0901& 0.0914 & 0.0984 & 0.0487 & \underline{0.0994}$\pm$0.0028 & \textbf{0.1076}$\pm$0.0001 & 8.30\%\\
                             & NDCG@5 & 0.0130 & 0.0058 & 0.0145 & 0.0241& 0.0261 & 0.0216 & 0.0115 & \underline{0.0281}$\pm$0.0018 & \textbf{0.0324}$\pm$0.0017 & 15.51\%\\
                            %  & NDCG@10& 0.0068 & 0.0181 & 0.0075 & 0.0226 & 0.0305 & 0.0300 & 0.0153 & \underline{0.0373} & \textbf{0.0393} & 5.36\%\\
                             & NDCG@20 & 0.0236 & 0.0104 & 0.0298 & 0.0387 & 0.0403 & 0.0391 & 0.0198 & \underline{0.0441}$\pm$0.0018 & \textbf{0.0489}$\pm$0.0013 & 10.90\%\\
    \midrule
    \multirow{4}{*}{Toys}  
    & HR@5  & 0.0120 & 0.0097 & 0.0166 & 0.0463 & 0.0502 & 0.0274 & 0.0143 & \underline{0.0526}$\pm$0.0034  & \textbf{0.0590}$\pm$0.0012 & 12.07\% \\
                            %  & HR@10  & 0.0113 & - & 0.0176 & 0.0270 & 0.0675 & 0.0450 & 0.0094 & \underline{0.0776} & \textbf{0.0830} & 6.96\% \\
                             & HR@20  & 0.0312 & 0.0301 & 0.0420 & 0.0941 & 0.0975 & 0.0688 & 0.0235 & \underline{0.1038}$\pm$0.0041  & \textbf{0.1150}$\pm$0.0016 & 10.74\%\\
                             & NDCG@5 & 0.0082 & 0.0059 & 0.0107 & 0.0306& 0.0337 & 0.0174 & 0.0123 & \underline{0.0362}$\pm$0.0025 &      
                        \textbf{0.0403}$\pm$0.0002 & 11.34\%\\
                            %  & NDCG@10 & 0.0163 & - & 0.0084 & 0.0141 & 0.0374 & 0.0231 & 0.0391 & \underline{0.0428} & \textbf{0.0479} & 11.92\%\\
                             & NDCG@20 & 0.0136 & 0.0116 & 0.0179 & 0.0441 & 0.0471 & 0.0291 & 0.0162 & \underline{0.0506}$\pm$0.0025 & 
                        \textbf{0.0560}$\pm$0.0004 & 10.57\%\\
    \midrule
    \multirow{4}{*}{Yelp}  
    & HR@5    & 0.0127 & 0.0152 & 0.0142 & 0.0160 & 0.0171 & 0.0196 & 0.0101 & \underline{0.0229}$\pm$0.0003  & \textbf{0.0257}$\pm$0.0007 & 12.23\%\\
                            %  & HR@10  & 0.0099 & 0.0216 & 0.0248 & 0.0254 & 0.0260 & 0.0339 & 0.0176 & \underline{0.0392} & \textbf{0.0426} & 8.67\% \\
                             & HR@20  & 0.0346 & 0.0371 & 0.0406 & 0.0443 & 0.0464 & 0.0564 & 0.0314 & \underline{0.0630}$\pm$0.0009 & \textbf{0.0677}$\pm$0.0016 & 7.47\%\\
                             & NDCG@5 & 0.0082 & 0.0091 & 0.008 & 0.0101 & 0.0112 & 0.0121 & 0.0068 & \underline{0.0144}$\pm$0.0001 & \textbf{0.0162}$\pm$0.0003 & 12.50\%\\
                            %  & NDCG@10 & 0.0051 & 0.0111 & 0.0124 & 0.0113 & 0.0133 & 0.0167 & 0.0092 & \underline{0.0197} & \textbf{0.0217} & 10.15\%\\
                             & NDCG@20 & 0.0143 & 0.0145 & 0.0156 & 0.0179 & 0.0193 & 0.0223 & 0.0127 & \underline{0.0256}$\pm$0.0003 & \textbf{0.0279}$\pm$0.0006 & 8.98\%\\
                             
  \bottomrule
\end{tabular}}
\end{table*}

\subsection{Experimental Setting}

% \begin{table}[htb]
% % \setlength{\abovecaptionskip}{0.01cm} 
% %   \setlength{\belowcaptionskip}{0cm} 
%   \caption{Dataset information.}
%   \label{tab:dataset-information}
%   \setlength{\tabcolsep}{2.0mm}{
%   \begin{tabular}{c|cccc}
%     \toprule 
%     Dataset   & Sports & Beauty & Toys & Yelp\\
%     \hline
%     $|\mathcal{U}|$ & 35,598 & 22,363 & 19,412 & 30,431\\
%     $|\mathcal{V}|$ & 18,357 & 12,101 & 11,924 & 20,033\\
%     \# Actions      & 0.3m   & 0.2m  & 0.17m & 0.3m\\
%     Avg. length     & 8.3    & 8.9   & 8.6  & 8.3\\
%     Sparsity        & 99.95\%& 99.95\%& 99.93\%& 99.95\%\\
%   \bottomrule
% \end{tabular}}
% \end{table}

\subsubsection{\textbf{Datasets}}
We conduct experiments on four public datasets. \emph{Sports}, \emph{Beauty} and \emph{Toys} 
%datasets 
are three subcategories of Amazon review data
%sets obtained from Amazon and 
introduced in~\cite{mcauley2015image}. Yelp\footnote{https://www.yelp.com/dataset} is a dataset for business recommendation. 
% The Steam dataset contains reviews from Steam video game platform and is introduced in~\cite{kang2018self}. 

We follow~\cite{zhou2020s3,xie2020contrastive} to prepare the datasets. In detail, we only keep the `5-core' datasets, in which all users and items have at least 5 interactions. The statistics of the prepared datasets are summarized in
Appendix~\ref{sec:dataset-info}.
% Table~\ref{tab:dataset-information}. 
% The users' interaction frequency in these datasets follow the long-tail distributions where most sequence.

\subsubsection{\textbf{Evaluation Metrics}} 
We follow~\cite{wang2019neural,krichene2020sampled} to rank the prediction on the whole item set without negative sampling. 
% All 
%the 
% Items are first ranked and then
%The p
Performance is
evaluated on 
a variety of evaluation metrics, including \textit{Hit Ratio}$@k$ ($\mathrm{HR}@k$), and \textit{Normalized Discounted 
Cumulative Gain}$@k$ ($\mathrm{NDCG}@k$) where $k\in\{5, 20\}$.

\subsubsection{\textbf{Baseline Methods}}
Four groups of 
baseline methods are included for comparison. 

\begin{itemize}
    \item \textbf{Non-sequential models:}
    % PopRec provides recommendations
    % based on the popularity of items;
    BPR-MF~\cite{rendle2012bpr} characterizes 
    the pairwise interactions
    via a matrix factorization model and
    optimizes through a
    pair-wise 
    Bayesian Personalized Ranking loss.
    \item \textbf{Standard sequential models.} 
    We include solutions
    that train the models with a next-item
    prediction objective.
    Caser~\cite{tang2018personalized} is a CNN-based approach, GRU4Rec~\cite{hidasi2015session} 
    is an RNN-based method, and SASRec~\cite{kang2018self} is one of the state-of-the-art Transformer-based
    baselines for SR. 
    \item \textbf{Sequential models with additional SSL:}
    BERT4Rec~\cite{sun2019bert4rec} 
    replaces the next-item prediction with 
    a \emph{Cloze} task~\cite{taylor1953cloze}
    to 
    fuse information 
    between an item (a view) in a user
    behavior sequence and its contextual information.
    S$^3\text{-Rec}$~\cite{zhou2020s3} uses SSL to capture correlation-ship among item, sub-sequence, and associated attributes from the given user behavior sequence.
    Its modules for mining on attributes are removed 
    because we don't have attributes for items,
    namely S$^3\text{-Rec}_{ISP}$.
    CL4SRec~\cite{xie2020contrastive} fuses
    contrastive SSL with a
    % self-attention
    Transformer-based SR model.
    \item \textbf{Sequential models considering latent factors}: 
    % Few work directly models the latent factor
    % on user behavior sequences.
    We include DSSRec\cite{ma2020disentangled},
    which utilizes seq2seq training 
    and performs optimization in
    latent space.
    We do not directly compare ASLI~\cite{tanjim2020attentive},
    as it requires user action type information (e.g.,
    click, add-to-favorite, etc).
    Instead, we provide a case study in
    Sec.~\ref{subsec:case_study} to
    evaluate the benefits of the learnt intent factor
    with additional item category information.
\end{itemize}

\subsubsection{\textbf{Implementation Details}}

Caser\footnote{https://github.com/graytowne/caser\_pytorch},
BERT4Rec\footnote{https://github.com/FeiSun/BERT4Rec},
and S3-Rec\footnote{https://github.com/RUCAIBox/CIKM2020-S3Rec}
are provided by the authors. 
BPRMF\footnote{https://github.com/xiangwang1223/neural\_graph\_collaborative\_filtering}, GRU4Rec\footnote{{https://github.com/slientGe/Sequential\_Recommendation\_Tensorflow}},
and 
DSSRec~\footnote{https://github.com/abinashsinha330/DSSRec}
are implemented based on public resources.
We implement SASRec and CL4SRec in PyTorch. 
The mask ratio in BERT4Rec is tuned
from $\{0.2, 0.4, 0.6, 0.8\}$. 
The number of attention heads and number of self-attention
layers for all self-attention based methods (SASRec, S$^3\text{-Rec}$, CL4SRec, DSSRec)
are tuned from $\{1, 2, 4\}$,
and $\{1, 2, 3\}$, respectively.
The number of latent factors introduced in DSSRec is tuned
from $\{1, 2,\dots, 8\}$.

% All other hyper-parameters 
% and initialization strategies are set based on 
%  suggestions from the methods' authors.

Our method is implemented in PyTorch.
Faiss~\footnote{https://github.com/facebookresearch/faiss} 
is used for $K$-means clustering
to speed up the training and query stages.
For the encoder architecture, 
we set self-attention blocks and attention heads as 2,
the dimension of the embedding as 64, 
and the maximum sequence length as 50.
The model is optimized by an Adam optimizer~\cite{kingma2014adam} with a
learning rate of 0.001, $\beta_{1}=0.9$, $\beta_{2}=0.999$, and
batch size of 256.
For hyper-parameters of \modelwithtansformer,
we tune $K$, $\lambda$ and $\beta$ within 
%the range of 
$\{8, 64, 128, 256,512,1024, 2048\}$, 
$\{0.1, 0.2,\cdots, 0.8\}$,
and $\{0.1, 0.2,\cdots, 0.8\}$ 
respectively. 
All 
%the
experiments are run
on a single Tesla V100 GPU.

\subsection{Performance Comparison}
\label{subsec:results}
Table~\ref{tab:main-results} shows
the results of different methods 
on all datasets. We have the following observations.
First, BPR performs worse 
than sequential models in general,
which indicates the importance of mining
the sequential patterns under user behavior sequences.
As for standard sequential models,
SASRec 
utilizes a Transformer-based encoder and 
achieves better performance than Caser and GRU4Rec.
This demonstrates the effectiveness of 
Transformer for capturing sequential patterns.
DSSRec further improves SASRec's
performance by
using a seq2seq training
strategy 
and reconstructs
the representation of 
the future sequence
in latent space for alleviating
non-convergence problems.

Moreover, 
though BERT4Rec and S$^3\text{-Rec}$ and adopt SSL to provide additional training
signals to enhance representations 
, we observe that both of them
exhibit worse performance
than SASRec in some datasets (e.g.,~in the Toys dataset).
The reason might be 
%because 
that both
BERT4Rec and S$^3\text{-Rec}$ 
%are
%aiming 
aim
to incorporate context information 
of  given user behavior sequences
via masked item prediction.
Such a goal may not align well with
the next item prediction target, and it
requires that each user behavior sequence is
long enough 
to provide comprehensive `context' information.
Thus their performances are degenerated when 
most sequences are short. 
Besides,
S$^3\text{-Rec}$ is proposed to fuse
additional contextual information. Without
such features, its two-stage training
strategy prevents 
%the 
information
sharing between the next-item prediction
and SSL tasks, thus leading to poor results. 
CL4SRec consistently performs
better than other baselines, 
demonstrating the effectiveness of
enhancing sequence representations via
contrastive SSL on an individual user level.

Finally, \modelwithtansformer consistently
outperforms existing methods on all datasets. 
The average improvements compared with the best baseline
%are ranging 
ranges
from \textbf{7.47\%} to \textbf{33.33\%} in $\mathrm{HR}$
and $\mathrm{NDCG}$. 
The proposed \lpmname 
estimates a good distribution of intents and fuses them
into SR model by a new contrastive SSL, 
which helps the encoder 
discover a good semantic structure across 
different user behavior sequences.

\begin{figure}
\begin{subfigure}[b]{\linewidth}
  \centering
  \includegraphics[width=0.95\linewidth]{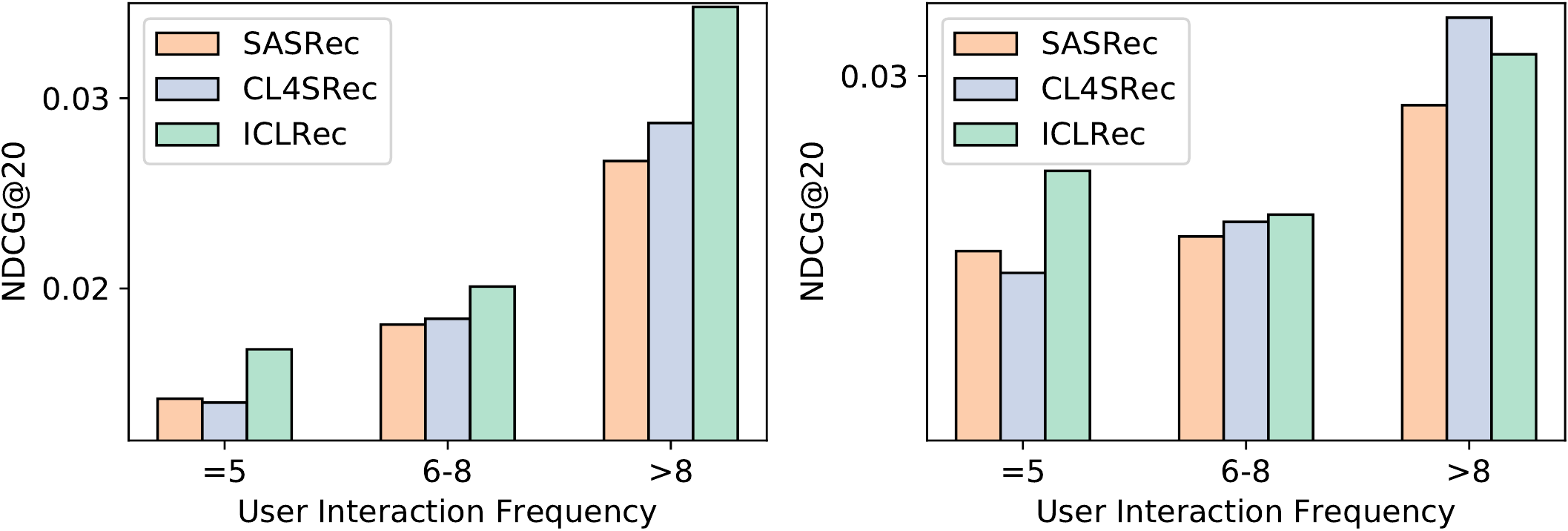}
%   \caption{Performance comparison on different user groups (Beauty and Yelp) among SASRec, CL4SRec and \modelwithtansformer.}
  \label{fig:robust-to-long-tail}
\end{subfigure}
\begin{subfigure}[b]{\linewidth}
  \centering
  \includegraphics[width=0.95\linewidth]{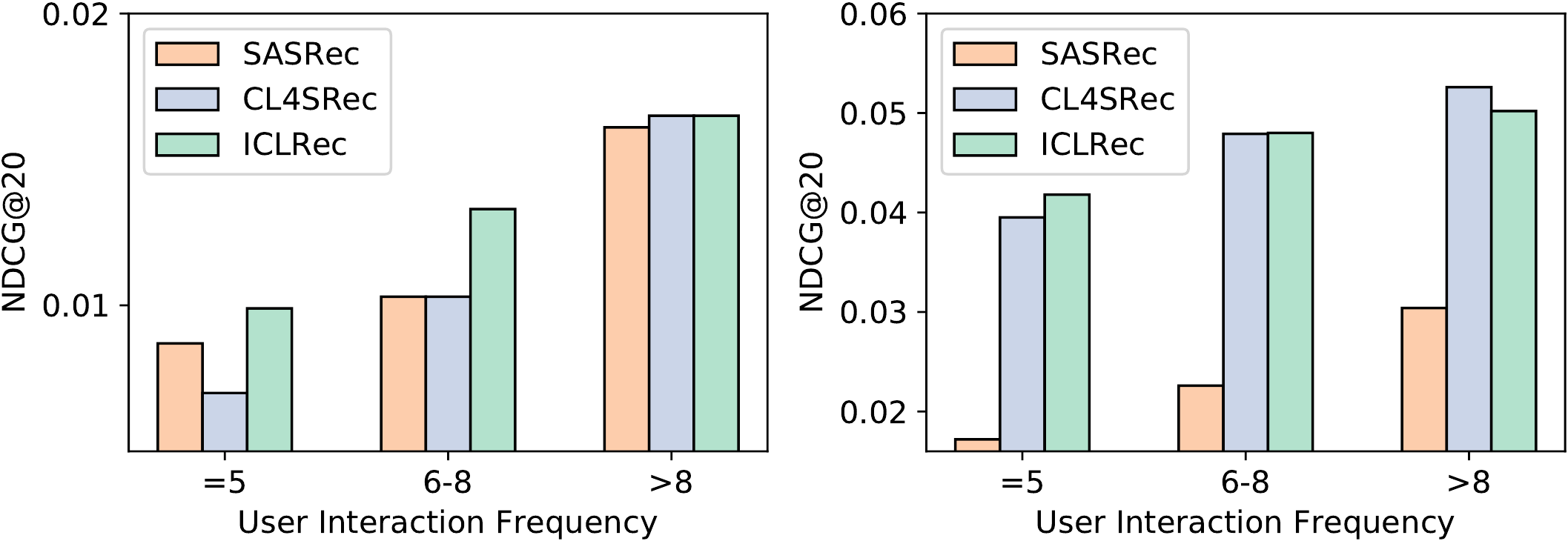}
%   \caption{Performance comparison on different user groups among SASRec, CL4SRec and \modelwithtansformer.}
  \label{fig:robust-to-long-tail-additional}
\end{subfigure}
\caption{Performance comparison on different user groups among SASRec, CL4SRec and \modelwithtansformer (Upper left: Beauty, Upper right: Yelp, Lower left: Sports, Lower right: Toys).)
}
\label{fig:robust-to-long-tail-all}
\end{figure}

% \begin{figure}[htb]
%   \centering
%   \includegraphics[width=0.9\linewidth]{./figures/robust_long_tail.pdf}
%   \caption{Performance comparison on different user groups (Beauty and Yelp) among SASRec, CL4SRec and \modelwithtansformer.}
%   \label{fig:robust-to-long-tail}
% \end{figure}

% \begin{figure}[htb]
%   \centering
%   \includegraphics[width=0.9\linewidth]{./figures/additional_robust_long_tail.pdf}
%   \caption{Performance comparison on different user groups among SASRec, CL4SRec and \modelwithtansformer.}
%   \label{fig:robust-to-long-tail-additional}
% \end{figure}

We also report the model efficiency
on Sports. SASRec is the most efficient solution.
It spends 3.59 s/epoch on model updates.
CL4SRec and the proposed \modelwithtansformer spend
6.52 and 11.75 s/epoch, respectively.
In detail, \modelwithtansformer spends 3.21 seconds
to perform intent representation learning, and rest of 8.54 
seconds
on 
multi-task learning.
The evaluation times of SASRec, CL4SRec, and \modelwithtansformer are about
the same($\sim$12.72s over testset) since 
the introduced \lpmname task 
%only
%be used on 
is only used during the
training stage. 

\subsection{\textbf{Robustness Analysis}}

\textbf{Robustness w.r.t.~user interaction frequency.}
\label{sec:user-interact-frequency}
The
user `cold-start' problem~\cite{cai2021category,yin2020learning} is one of the
typical data-sparsity issues that
recommender systems often face, i.e.,
most users have limited historical behaviors.
To check whether \lpmname improves the robustness
under such a scenario,
%or not, 
we
split user behavior sequences into three groups
based on their behavior sequences' length, and keep
the total number of behavior sequences 
%are 
the same.
Models are trained and evaluated on each group of users
independently. Figure~\ref{fig:robust-to-long-tail-all}
shows the comparison results on four datasets. 
We observe that:
(1) The proposed \modelwithtansformer can
consistently 
performs better than
SASRec among all user groups while
CL4SRec fails to outperform SASRec
in Beauty and Yelp
when user behavior sequences are
short. 
This demonstrates that
CL4SRec requires individual user 
behavior sequences long enough to provide
`complete' information 
for auxiliary supervision
while
\modelwithtansformer reduces
the need by leveraging 
user intent information, 
% which can been seen as collaborative signal,
thus consistently benefiting user
representation learning even
when users have limited historical interactions.
(2) Compared with CL4SRec,
we observe that
the improvement of \modelwithtansformer
is mainly because 
it provides better
recommendations to 
users with low interaction frequency. 
This
verifies that
user intent information
is beneficial, especially
when the recommender system faces
data-sparsity issues where 
information
in each individual user sequence 
is limited.

\begin{figure}[htb]
  \centering
  \includegraphics[width=0.95\linewidth]{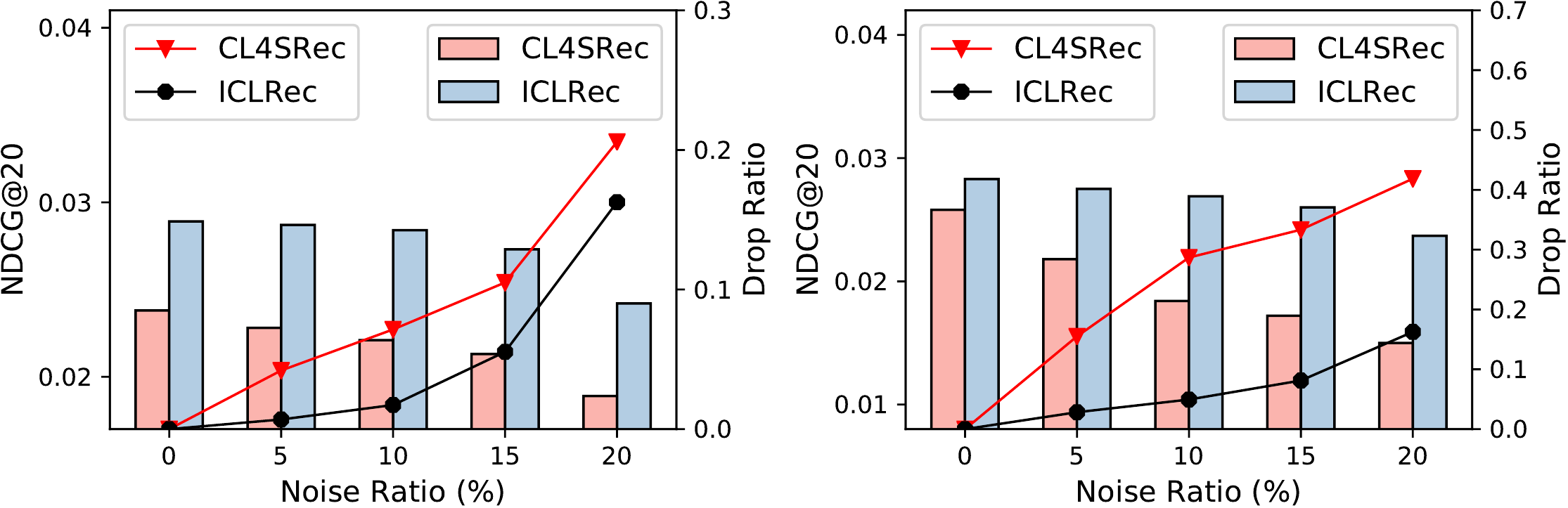}
  \caption{Performance comparison w.r.t.~noise ratio on Sports and Yelp.
  The bar chart shows the performance in NDCG@5 and the line chart 
  shows the corresponding drop rate.}
  \label{fig:robust-to-noise}
\end{figure}

\textbf{Robustness to Noisy Data.}
We also conduct experiments on the Sports
and Yelp datasets
to verify the robustness of
\modelwithtansformer against 
noisy interactions in the test phase.
Specifically, 
we randomly add 
a certain proportion 
(i.e., 5\%, 10\%, 15\%, 20\%) of negative items 
to text sequences. 
From Figure~\ref{fig:robust-to-noise}
we can see that adding noisy data deteriorates
the performance of CL4SRec and \modelwithtansformer.
However, the performance drop ratio of 
\modelwithtansformer is consistently
lower than CL4SRec, and its performance
with 15\% noise proportion 
can still outperforms  CL4SRec
without noisy dataset on Sports.
The reason might be
the leveraged intent information
is collaborative information that
distilled from all the users. 
\lpmname helps the SR model
capture semantic structures
from user behavior sequences, 
which increases the robustness of \modelwithtansformer to noisy perturbations on individual sequences.

% \subsection{Effect of ICL with different Local MIM (done)}
% \label{sec:effect-local-signal}
% Comparison between ICL+CL4SRec, and ICL+S3SRec.

% \subsection{\textbf{Effect of False Negative Noise Cancellation}}
% TODO

\begin{table}[htb]
\caption{Ablation study of \modelwithtansformer (NDCG@20). }
\label{tab:ablation-study}
\begin{tabular}{l|cccc}
\toprule
\multirow{2}{*}{Model} & \multicolumn{4}{c}{Dataset} \\
                       & Sports & Beauty & Toys & Yelp \\
                    %   & NDCG@20 & NDCG@20 & NDCG@20 & NDCG@20 \\ 
\hline
(A) \modelwithtansformer  & \textbf{0.0287} & \textbf{0.0480} & \textbf{0.0554} & \textbf{0.0283} \\
% \hline
(B) w/o FNM       & 0.0283 & 0.0465 & 0.0524 & 0.0266 \\
(C) only \lpmname & 0.0263 & 0.0429 & 0.0488 & 0.0267  \\
(D) w/o \lpmname  & 0.0238 & 0.0428 & 0.0505 & 0.0258  \\
(E), is (C) w/o seq. aug & 0.0242 & 0.0414 & 0.0488 & 0.0213 \\
(F) SASRec & 0.0216 & 0.0387 & 0.0441 & 0.0179 \\
\hline
(G) \lpmname + S$^3\text{-Rec}_{ISP}$              & \textbf{0.0157} & \textbf{0.0264} & \textbf{0.0266} & \textbf{0.0205}  \\
(H) S$^3\text{-Rec}_{ISP}$      & 0.0146 & 0.0198 & 0.0162 & 0.0127  \\
% \hline
% (G) \lpmname+CASER             & - & - & - & -  \\
% (H) CASER   & 0.0179 & 0.0298 & 0.0178 & 0.0156  \\
% \hline

\bottomrule
\end{tabular}
\end{table}

\subsection{Ablation Study}
\label{sec:ablation}
Our proposed \modelwithtansformer
contains a novel \lpmname objective,
a false-negative noise mitigation (FNM) 
strategy,
a SeqCL objective,
and sequence augmentations. 
To verify the effectiveness 
of each component,
we conduct an ablation study 
on four datasets and report results 
in Table~\ref{tab:ablation-study}.
(A) is our final model, and (B) to (F)
are \modelwithtansformer removed certain components.
From (A)-(B) we can see that
the FNM leverages the learned intent information
to avoid users with similar intents
pushing away in their representation space
which helps the model to learn better user representations.
Compared with (A)-(D), we find that
without
the proposed
\lpmname, the performance
drops significantly, which demonstrates
the effectiveness of \lpmname.
Compared with (A)-(C), we find that 
individual user level mutual information
also 
%benefits for enhancing
helps to enhance
user representations. As we analyze in 
Sec.~\ref{sec:user-interact-frequency},
it contributes more to long user sequences. 
Compared with (E)-(F), we find that
\lpmname can 
%performance 
perform
contrastive SSL
without sequence augmentations and outperforms
SASRec. While CL4SRec requires the
sequence augmentation module to perform
contrastive SSL. Comparison between (C) and (E)
indicates sequence augmentation enlarges training 
set, which benefits improving performance.

Since \lpmname is a model-agostic learning
paradigm, we also add \lpmname
to the S$^3\text{-Rec}_{ISP}$~\cite{zhou2020s3} 
model in the fine-tuning stage
to further verify its effectiveness.
Results are shown in 
Table.~\ref{tab:ablation-study} (G)-(H). We find that
the S$^3\text{-Rec}_{ISP}$ model 
also benefits from the \lpmname
objective. The average
improvement over 
the four datasets is
41.11\% in NDCG@20,
which further validate
the effectiveness and 
practicality of \modelwithtansformer.
% we also replace the user level SSL
% with an item level SSL objective proposed by
% S$^3\text{-Rec}_{ISP}$ and shown in (E).
% However,

% Ablation studies w.r.t. intent contrastive learning objective, false negative noise mitigate, and two different sequence level mutual signals.

\subsection{Hyper-parameter Sensitivity}

\begin{figure}[htb]
  \centering
  \includegraphics[width=\linewidth]{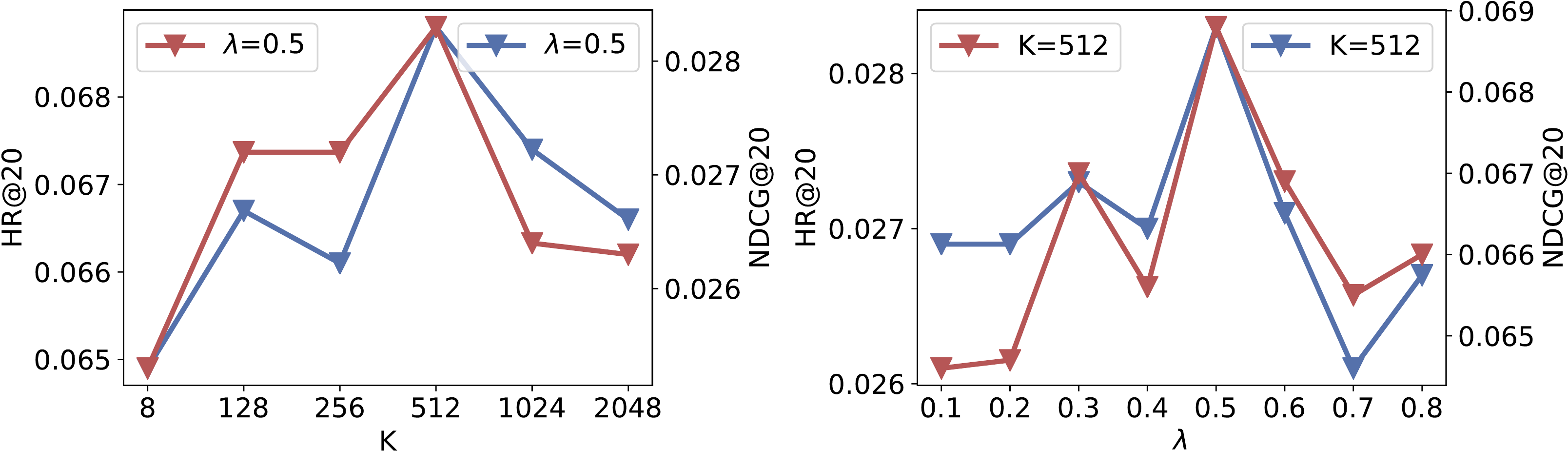}
  \caption{Impact of intent class numbers $K$ and the intent contrastive learning strength $\lambda$ on Yelp.}
  \label{fig:hpo-study}
\end{figure}

The larger of the intent class number $K$ means users can have
more diverse 
intentions.
The larger value of the strength of SeqCL objective $\beta$
means the \lpmname task contributes more to 
the final model.
The results on Yelp is shown in Figure~\ref{fig:hpo-study}.
We find that: (1) 
\modelwithtansformer reaches its 
best performance when increasing $K$ to 512,
and then it starts to deteriorate
as $K$ become larger. 
When $K$ is very small,
the number of users under
each intent prototype can potentially be large.
% most users in a training batch 
% are assigned with similar intent prototypes.
% The imperfect user intent assignment 
As a result, false-positive samples 
(i.e., users that actually have different intents
are considered as having the same intent erroneously)
are introduced to the contrastive SSL,
thus affecting learning.
On the other hand, when $K$ is too large,
the number of users under
each intent prototype is small,
the introduced false-negative samples
will also impair contrastive SSL.
In Yelp, 512 user intents summarize
users' distinct behaviors best.
(2) A `sweet-spot' of $\lambda=0.5$ can 
also be found. 
It indicates
that the \lpmname task can benefit
the recommendation prediction
as an auxiliary task.
The impact of the batch size and $\beta$ are provided in Appendix~\ref{sec:additional-hpo}.

\begin{figure}[htb]
  \centering
  \includegraphics[width=1.0\linewidth]{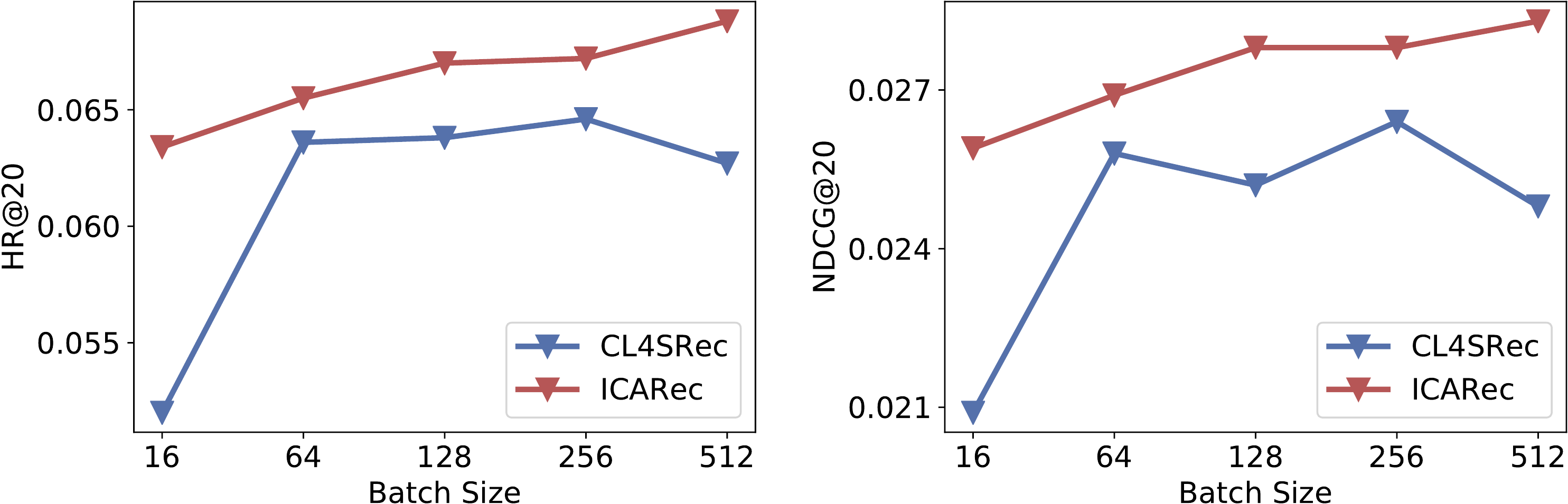}
  \caption{Performance comparison w.r.t.~Batch Size.}
  \label{fig:batch_size}
\end{figure}

% \subsubsection{\textbf{Effect of Intent Cluster Number}}
% TODO. Quick observation from result tables are 32-256 are general good intent size in amazon datasets and yelp, more results on lastfm and toys to be done.

% \subsubsection{\textbf{Comparison of Mutual Information on Sequence Level and Intent Level}}
% TODO. Objective weights hyper-parameter study. Observation from current results are, SeqCL weight at 0.1 while IntentCL at 0.3 - 0.5 gives best performances, maybe show that it is more important to capture mutual information about common/structure intent information, the low-level/sequence level mutual information which consider as detailed information as complementary are relatively less important.

\subsection{Case Study}
\label{subsec:case_study}

The Sports dataset~\cite{mcauley2015image} 
contains 2,277 fine-grained item categories,
and the Yelp dataset provides 1,001
business categories.
We utilize these attributes
to study the effectiveness of the proposed 
\modelwithtansformer both
quantitatively and qualitatively.
Note that we did not use this information during 
the training phrase. 
The detailed analysis results are in Appendix~\ref{subsec:case_study-appendix}.

\section{Conclusion}

In this work,
we propose a new learning paradigm \lpmname
that can
model latent intent factors
from user interactions
%behaviors,
and fuse them into a sequential recommendation
model via a new
contrastive SSL objective.
\lpmname is formulated within
an EM framework, which guarantees
convergence. Detailed analyses show
the superiority of \lpmname and 
experiments conducted on
four datasets further demonstrate 
the effectiveness of the proposed method.
% In the future, we will consider 
% exploring advanced  `Aggregation' layer,
% reduce training computation cost of \lpmname
% by implementing a parallel training pipeline,
% and improve the E-step of \lpmname, such as replacing
% K-means with KNN, etc.

\newpage
\bibliographystyle{ACM-Reference-Format}
\bibliography{sample}
% \newpage
\appendix

\section{Pseudo-code of \lpmname for SR}
\label{sec：algorithm}
\begin{algorithm}[htb]
\SetKwInput{KwInput}{Input}                
\SetKwInput{KwOutput}{Output} 
\KwInput{training dataset $\left \{ s_{u} \right\}_{u=1}^{|\mathcal{U}|}$, sequence encoder $f_{\theta}$, batch size $N$ , hyper-parameters $K$, $\lambda$, $\beta$.}
\KwOutput{$\theta$.}
 \While{ $epoch \leq MaxTrainEpoch$}{
    \tcp{E-step: Intent Representation Learning}
    % $\left \{ \mathbf{h}_{u} \right\}_{u=1}^{|\mathcal{U}|} = \left \{ f_{\theta}(s_{u}) \right\}_{u=1}^{|\mathcal{U}|}$\\
    $c = Clustering(\left \{ f_{\theta}(S^{u}) \right\}_{u=1}^{|\mathcal{U}|}, K)$\\
    Update distribution function $Q(c_{i}) = P_{\theta}(c_{i}|S^{u})$
    
    \tcp{M-step: Multi-Task Learning}
    \For{a minibatch $\{s_{u}\}_{u=1}^{N}$}{
        \For{$u \in \{1, 2, \cdots, N\}$}{
              \tcp{Construct $2$ views.}
                $\Tilde{S}^{u}_{1} = g_{1}^{u}(S^{u}), \Tilde{S}^{u}_{2} = g_{2}^{u}(S^{u}), ~where~g_{1}^{u}, g_{2}^{u} \sim \mathcal{G}$
              
              \tcp{Encoding via $f_{\theta}(\cdot)$} 
              
              $\mathbf{h}^{u} = f_{\theta}(S^{u})$ \\
              $\mathbf{\Tilde{h}}^{u}_{1}  =   f_{\theta}(\Tilde{S}^{u}_{1}) ,~\mathbf{\Tilde{h}}^{u}_{2}  =   f_{\theta}(\Tilde{S}^{u}_{2})$
            %   $\tilde{\mathbf{h}}_{2u-1},\tilde{\mathbf{h}}_{2u} = f_{\theta}(\tilde{s}_{2u-1}),f_{\theta}(\tilde{s}_{2u})$
        }
         
        \tcp{Optimization}
          $\mathcal{L}= \mathcal{L}_{\text{NextItem}} + \lambda \cdot \mathcal{L}_{\text{\lpmname}}
                + \beta \cdot \mathcal{L}_{\text{SeqCL}}
          $ \\
          Update network $f_{\theta}(\cdot)$ to minimize $\mathcal{L}$
    }
 }

 \caption{\lpmname for SR}
 \label{alg:icl}
\end{algorithm}

\begin{figure}[htb]
  \centering
  \includegraphics[width=.9\linewidth]{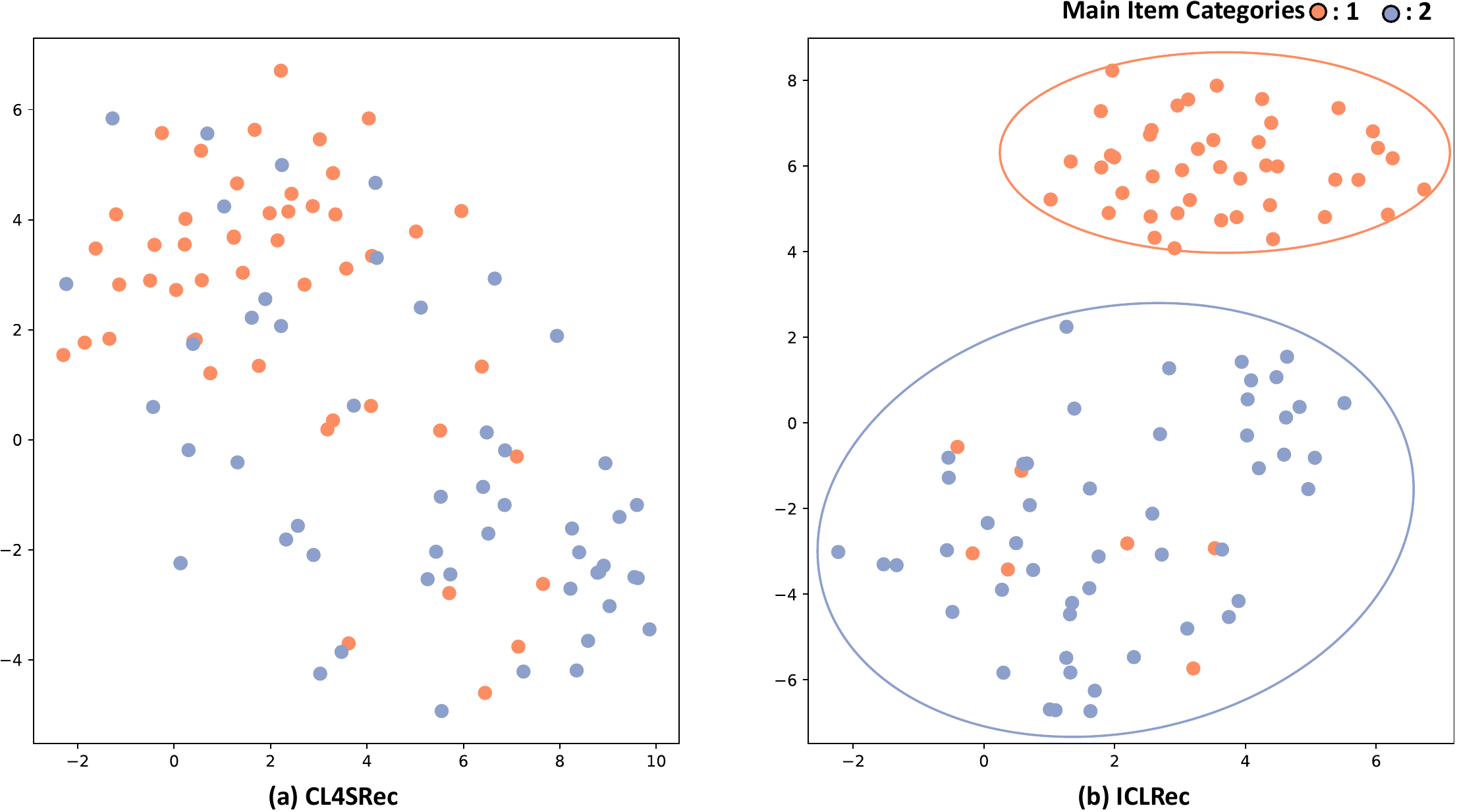}
  \caption{Visualization of the learned users' representations by CL4SRec and \modelwithtansformer on Sports.}
  \label{fig:case-study}
\end{figure}

% \section{Derivative of the Lower-bound}
% \label{sec:lower-bound}
\section{Proof of Convergence}
\label{sec:em-convergence}
To proof the 
convergence of \lpmname under the generalized EM framework,
we just need to proof $P_{\theta^{(m+1)}}(S)\geq P_{\theta^{(m)}}(S)$,
where $m$ indicates the
number of training iterations.
Based on Eq.~(\ref{eq:3}), we have 
\begin{equation}
\label{eq:conv-proof-1}
\begin{split}
\ln P_{\theta^{(m)}}(S) = \ln\frac{P_{\theta^{(m)}}(S, c_{i})}{P_{\theta^{(m)}}(c_{i}|S)}
                    = \ln P_{\theta^{(m)}}(S, c_{i}) - \ln P_{\theta^{(m)}}(c_{i}|S).
\end{split}
\end{equation}

Take the expectation in term of $c$ condition over $S$ on both sides, then we have:

\begin{equation}
\label{eq:conv-proof-2}
\begin{split}
\mathbb{E}_{(c|S,\theta^{(m)})} \left[\ln P_{\theta}(S)\right] &=  \mathbb{E}_{(c|S,\theta^{(m)})}\left[\ln P_{\theta^{(m)}}(S, c)\right] \\
        &- \mathbb{E}_{(c|S,\theta^{(m)})}\left[\ln P_{\theta_{(m)}}(c|S)\right].
\end{split}
\end{equation}

Based on Eq.~(\ref{eq:q-distribution}), and~\ref{eq:conv-proof-2}, the term on left side equal to:
\begin{equation}
\label{eq:conv-proof-3}
\begin{split}
\mathbb{E}_{(c|S,\theta^{(m)})} \left[\ln P_{\theta^{(m)}}(S)\right] &= 
                    \sum_{i=1}^{K}
                    Q(c_{i})\cdot\ln P_{\theta^{(m)}}(S)
                    = \ln P_{\theta^{(m)}}(S).
\end{split}
\end{equation}
Thus, proof $P_{\theta^{(m+1)}}(S)\geq P_{\theta^{(m)}}(S)$
is equivalent to proof 
\begin{equation}
\label{eq:conv-proof-4}
\begin{split}
\ln P_{\theta^{(m+1)}}(S) \geq \ln P_{\theta^{(m)}}(S),
\end{split}
\end{equation}
which is equivalent to:
\begin{equation}
\label{eq:conv-proof-5}
\begin{split}
\mathbb{E}_{(c|S,\theta^{(m+1)})}\left[\ln P_{\theta^{(m+1)}}(S, c)\right] - \mathbb{E}_{(c|S,\theta^{(m+1)})}\left[\ln P_{\theta_{(m+1)}}(c|S)\right] \\
\geq \mathbb{E}_{(c|S,\theta^{(m)})}\left[\ln P_{\theta^{(m)}}(S, c)\right] - \mathbb{E}_{(c|S,\theta^{(m)})}\left[\ln P_{\theta_{(m)}}(c|S)\right].
\end{split}
\end{equation}

Because we try to optimize $\theta$ at M-step,
thus we have
\begin{equation}
\label{eq:conv-proof-6}
\begin{split}
\mathbb{E}_{(c|S,\theta^{(m+1)})}\left[\ln P_{\theta^{(m+1)}}(S, c)\right]
\geq \mathbb{E}_{(c|S,\theta^{(m)})}\left[\ln P_{\theta^{(m)}}(S, c)\right].
\end{split}
\end{equation}

And based on Jsnson’s inequality, we 
have

\begin{equation}
\label{eq:conv-proof-7}
\begin{split}
\mathbb{E}_{(c|S,\theta^{(m+1)})}\left[\ln P_{\theta_{(m+1)}}(c|S)\right]
\leq \mathbb{E}_{(c|S,\theta^{(m)})}\left[\ln P_{\theta_{(m)}}(c|S)\right].
\end{split}
\end{equation}

Combining Eq.~(\ref{eq:conv-proof-5}),~(\ref{eq:conv-proof-6}),and~(\ref{eq:conv-proof-7}), we show that $P_{\theta^{(m+1)}}(S)\geq P_{\theta^{(m)}}(S)$.
Thus, the algorithm will converge.

\section{Dataset Information}
\label{sec:dataset-info}
\begin{table}[htb]
  \caption{Dataset information.}
  \label{tab:dataset-information}
  \setlength{\tabcolsep}{2.0mm}{
  \begin{tabular}{c|cccc}
    \toprule 
    Dataset   & Sports & Beauty & Toys & Yelp\\
    \hline
    $|\mathcal{U}|$ & 35,598 & 22,363 & 19,412 & 30,431\\
    $|\mathcal{V}|$ & 18,357 & 12,101 & 11,924 & 20,033\\
    \# Actions      & 0.3m   & 0.2m  & 0.17m & 0.3m\\
    Avg. length     & 8.3    & 8.9   & 8.6  & 8.3\\
    Sparsity        & 99.95\%& 99.95\%& 99.93\%& 99.95\%\\
  \bottomrule
\end{tabular}}
\end{table}

\section{Impact of Batch Size and the strength of SeqCL task $\beta$}
\label{sec:additional-hpo}

\begin{figure}[htb]
  \centering
  \includegraphics[width=1.0\linewidth]{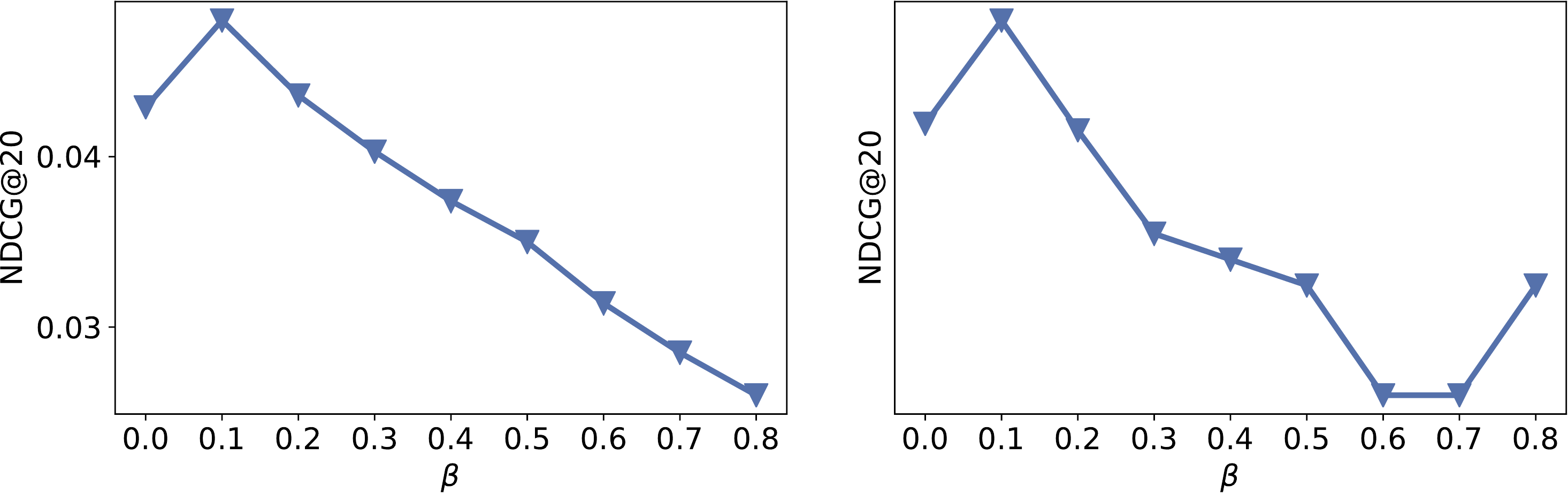}
  \caption{Impact of SeqCL task strength $\beta$ on Beauty (left) and Yelp (right).}
  \label{fig:beta}
\end{figure}

Performance w.r.t.~batch size on Yelp between CL4SRec and the proposed \modelwithtansformer
are shown in Figure.~\ref{fig:batch_size}.
% Since existing contrastive SSL for SR 
% highly depends on the 
% training batch size to create
% negative samples for contrastive loss. 
% Results are shown in Figure.~\ref{fig:batch_size}. 
We observe that with the batch size increases,
CL4SRec's performance does not continually improve.
The reason might because of larger batch sizes
introduce false-negative samples, which harms 
learning. While \modelwithtansformer
is relatively stable with different batch sizes,
and out performs CL4SRec in all circumstances. Because the intent
learnt can be
seen as a pseudo label of sequences,
which helps identify the true
positive samples
via the proposed contrastive SSL with FNM.

The impact of $\beta$ is shown in Figure~\ref{fig:beta}. 
We can see that,
$\beta$ does help
\modelwithtansformer improve the performance
when it is small (e.g., $\beta \leq 0.1$). 
However, when $\beta$ continually increase,
the model performance drop significantly.
This phenomenon also indicates the 
limitation of SeqCL, since focusing on 
maximize mutual
information between individual sequence pairs 
may break the global relationships among users.
% The main 
% observation
% is that $\beta$ only benefits model learning
% when it is small. The overall improvement of \modelwithtansformer are mainly from other components.

% Existing contrastive SSL for SR highly depending on the 
% training batch size to create
% negative samples for contrastive loss. 
% We provide comparison results on Yelp between CL4SRec and the proposed \modelwithtansformer  w.r.t.~batch size shown in Figure.~\ref{fig:batch_size}. 
% We observe that with the batch size increases,
% CL4SRec's performance do not continually improved.
% The reason might because that larger batch size
% introduces false-negative samples, which harms 
% learning. While \modelwithtansformer
% is relatively stable with different batch size,
% and out performances CL4SRec in our circumstance. Because the intent
% learnt in E-step can be
% seen as a pesudo label of sequences,
% which helps identifies the true
% positive and true negative samples
% via FNM strategy.

\begin{table}[htb]
\caption{Quantitative Analysis Results. (NDCG@20)}
\label{tab:case-study}
\begin{tabular}{c|cccl}
\toprule
\multirow{1}{*}{Datasets} 
                       & SASRec & CL4SRec & $\modelwithtansformer\text{-A}$ & \multicolumn{1}{c}{\modelwithtansformer}\\
                    %   & NDCG@20 & NDCG@20 & NDCG@20 & NDCG@20 \\ 
\hline
\multirow{2}{*}{Sports}    & \multirow{2}{*}{0.0216} & \multirow{2}{*}{0.0238} & \multirow{2}{*}{0.0272}  & 0.0275 ($K=2048$) \\
& & & & \textbf{0.0287} ($K=1024$) \\
\hline
\multirow{2}{*}{Yelp}     & \multirow{2}{*}{0.0179} & \multirow{2}{*}{0.0258} & \multirow{2}{*}{0.0264} & 0.0271 ($K=1024$)\\
& & & & \textbf{0.0283} ($K=512$) \\
\bottomrule
\end{tabular}
\end{table}

\section{Case Study}
\label{subsec:case_study-appendix}

% The Sports dataset~\cite{mcauley2015image} 
% contains 2,277 fine-grained item categories,
% and the Yelp dataset provides 1,001
% business categories.
% We utilize these attributes
% to study the effectiveness of the proposed 
% \modelwithtansformer both
% quantitatively and qualitatively.
% Note that we did not use this information during 
% the training phrase.

\textbf{Quantitative analysis}.
We study how \modelwithtansformer will perform
by considering the item categories of 
users interacted items as their intents. 
Specifically, 
given a user behavior sequence $S^{u}$,
we consider the mean of its corresponding
trainable item category embeddings as
the intent prototype $\mathbf{c}$,
aiming to replace the intent representation
learning 
%that
described in Sec.~\ref{sec:intent-represent-learning}.
We run the corresponding model named $\modelwithtansformer\text{-A}$
and
show the comparison results in Table~\ref{tab:case-study}.
We observe that on Sports (1) $\modelwithtansformer\text{-A}$ performs
better than CL4SRec, which shows the
potential benefits of leveraging item category information.
(2) \modelwithtansformer achieves similar performance 
as $\modelwithtansformer\text{-A}$'s when $K=2048$. Joint analysis with
the above qualitative results
%, 
%it 
indicates that \lpmname can capture meaningful user intents
via SSL. (3) \modelwithtansformer can outperform 
$\modelwithtansformer\text{-A}$ when $K=1024$. We hypothesize
that users' intents can be better described by the
latent variables when $K=1024$ thus improving performance.
(e.g.,~parents of the existing item categories.) 
Similar observations in Yelp.

\textbf{Qualitative analysis}
We also compare the proposed \modelwithtansformer with CL4SRec
by visualizing the learned users'
representations via t-SNE~\cite{van2008visualizing}. 
% Please see Appendix~\ref{sec:case-qualitative} for more details.
% \section{Qualitative analysis}.
% \label{sec:case-qualitative}
Specifically, we sampled 100 users for whom 
used to interact with one category of
items or the other category. 
These 100 users also interacted with other
categories of items in the past.
We visualize the learned users’ representations 
via t-SNE~\cite{van2008visualizing}
in Figure~\ref{fig:case-study}. 
From Figure~\ref{fig:case-study} we can see,
users' representations learned by \modelwithtansformer
intent to pull users 
that interacted with the same category of items
closer to each other while pushing others 
%more
further
away
in the representation space than CL4SRec. It reflects
that representations learned by \lpmname
can capture more semantic structures, 
therefore, improves the performance.

\end{document}